\documentclass[conference]{IEEEtran}
\IEEEoverridecommandlockouts

\usepackage{cite}
\usepackage{amsmath,amssymb,amsfonts}
\usepackage{algorithmic}
\usepackage{graphicx}
\usepackage{textcomp}
\usepackage{xcolor}
\usepackage{bbm}
\usepackage{colortbl} 
\usepackage{graphicx}
\usepackage{booktabs}
\usepackage{multirow}
\usepackage{makecell}
\usepackage{arydshln}
\usepackage{rotating}
\usepackage{setspace}
\usepackage{bbm}
\usepackage{subcaption}
\usepackage{pifont}
\usepackage[dvipsnames]{xcolor}
\usepackage{hyperref}

\def\BibTeX{{\rm B\kern-.05em{\sc i\kern-.025em b}\kern-.08em
    T\kern-.1667em\lower.7ex\hbox{E}\kern-.125emX}}
\begin{document}

\title{Localized Conformal Prediction for Image Classification with Vision-Language Models\\
}




\author{
\IEEEauthorblockN{Cl\'ement Fuchs*, Tim Bary*, Beno\^it Macq\thanks{*C.~Fuchs and T.~Bary contributed equally. C.~Fuchs and T.~Bary are funded by the MedReSyst project, FEDER and the Walloon Region. Computational resources have been provided by the CÉCI, funded by the F.R.S.-FNRS under Grant No. 2.5020.11 and the Walloon Region.}}
\IEEEauthorblockA{
\textit{ICTEAM, UCLouvain} \\
Louvain-la-Neuve, Belgium \\}}

\maketitle

\begin{abstract}
Conformal predictions have attracted significant attention in the field of uncertainty quantification, mainly because of their strong marginal coverage guarantees. Full conditional guarantee is not an attainable goal, a well known fact in conformal predictions literature. As a result, several approaches have tried to approximate this behavior by adapting the conformal sets of test-time samples according to their similarity to calibration examples. Although the latter has gained traction and shown impressive performances for regression problems, its application to image classification remains under-explored. We conduct an extensive benchmarking on natural image classification tasks with vision-language models (VLMs), using our open source implementation of a recent localized conformal prediction algorithm. We show that straightforward usage of the cosine similarity between test-time and calibration visual features, an intuitive choice for VLMs, is not sufficient to improve over the non-local baselines. In response, we propose a simple non-linear transformation of the cosine similarities, which conserves marginal coverage guarantees and achieves statistically significant mean set sizes reduction. Code is available at \href{https://github.com/cfuchs2023/lcp-vlm/}{github.com/cfuchs2023/lcp-vlm/}.

\end{abstract}

\section{Introduction}
Modern machine learning models achieve remarkable accuracy on visual tasks such as object detection, medical image analysis, and human action recognition. However, these models typically produce point predictions without quantifying uncertainty, which is a major shortcoming in safety-critical applications. In domains such as medical diagnostics, autonomous driving, or environmental monitoring, reliable uncertainty estimates help identify ambiguous inputs, flag unreliable predictions, and guide human oversight~\cite{straitouridesigning, cresswell2024conformal}.

Conformal prediction (CP)  is a statistical framework for uncertainty quantification in machine learning. Given a predictive model and a set of calibration data, CP produces prediction sets or intervals that guarantee a user-defined level of coverage, assuming exchangeability of the data---a weaker assumption than i.i.d \cite{vovk2005algorithmic}. Due to its model-agnostic nature and theoretical guarantees, CP has been adopted in areas where reliable uncertainty estimates are crucial, such as medical imaging and human-in-the-loop systems.

Despite these strengths, a well-known limitation of CP is that it only guarantees marginal coverage (\textit{i.e.}, averaged over the entire data distribution), but not conditional coverage for subgroups (\textit{e.g.}, specific classes or feature regions) \cite{angelopoulos2020uncertainty}. This leads to coverage disparities, where some subpopulations are under- or over-covered. These disparities often stem from the global nature of the calibration step, which assigns equal importance to all calibration examples.

Localized Conformal Prediction (LCP), introduced in \cite{guan2023localized}, addresses this limitation by weighting calibration examples based on their similarity to the test sample. This yields test-specific prediction intervals that adapt to local distributional properties. While LCP retains the marginal coverage guarantees of standard CP, it can also offer improved local coverage under suitable conditions. So far, however, LCP has mainly been studied in regression settings and tested on synthetic datasets.

In this paper, we apply the LCP framework to classification tasks and benchmark it against standard CP on an extensive collection of datasets. Our approach integrates LCP into Vision-Language Models (VLMs) by leveraging their latent representations to define similarity. We show that using a non-linear transformation of the cosine similarity to weight the calibration set leads to improved set sizes and class-conditional coverage compared to traditional CP.

Our contributions are threefold:
\begin{enumerate}
    \item \textbf{LCP for Classification:} We benchmark LCP for classification tasks using VLMs and evaluate it with different conformal classification scores such as LAC, APS, and RAPS. Additionally, we show that cosine similarity is a suboptimal weighting function for LCP in classification and propose a non-linear transformation to enhance its effectiveness.
    \item \textbf{Extensive Benchmarking:} We evaluate our method on nine diverse image classification datasets, showing that LCP significantly improves average prediction set size.
    \item \textbf{Open-Source Implementation:} We release the first, to our knowledge, open-source implementation of LCP for classification, compatible with a wide range of datasets and models.
\end{enumerate}

The remainder of this article is structured as follows.  Section~\ref{sec:background} reviews CP and LCP, including classification-specific conformal scores and an overview of VLMs. Section~\ref{sec:methods} describes our experimental setup and introduces the proposed non-linear transformation of cosine similarity. In Section~\ref{sec:results}, we report and analyze the experimental results. Finally, Section~\ref{sec:conclusion} concludes the paper and outlines directions for future work.

\section{Background}\label{sec:background}
\subsection{Conformal prediction}
Conformal predictions methods generally follow the same basic principle. First, conformal scores $s_i (k^{*}_i)$ are computed for $n$ calibration samples, which differ from the training, validation, and test samples, with the knowledge of their ground-truth classes $k_i^{*}$. Then, the $1-\alpha$ quantile of these scores are computed as 
\begin{equation}
q_{\alpha} = \text{Quantile}\left(\{s_i (k_i^{*}) ; 1 \leq i \leq n\} ; \dfrac{n+1}{n} (1 - \alpha)\right)
\end{equation}
 for a given error level $\alpha$. Finally, for a test time sample $x_{\text{test}}$, all classes $k$ such that $s_{\text{test}}(k) \leq q_{\alpha}$ are retained to form the conformal prediction set $C(\mathbf x_{\text{test}})$. One major difference between conformal prediction methods is the way they construct their conformal scores from the prediction vectors $y_i$, belonging to the $K-$simplex $\Delta_K$, given by the model. In the following, we present four different scores which are ubiquitous in the literature.

\paragraph{TopK} The TopK conformal methods uses the rank of the class in the prediction vector as the conformal scores. Formally, let $P_i$ the permutation function yielding the indexes sorting the components of $y_{i,k}$ in decreasing order. Then, the TopK conformal score reads as 
\begin{equation}
s^{\text{TopK}} (k^*_i) = P_i(k^*_i).
\end{equation}
\paragraph{Least Ambiguous set-valued Classifier} The Least Ambiguous set-valued Classifier~(LAC)~\cite{vovk2005algorithmic,sadinle2019least,lei2015conformal} uses conformal scores defined as follow : 
\begin{equation}\label{eq:nonconf}
s^{\text{LAC}}(k^{*}_i) = 1 - y_{i,k_i^*}.
\end{equation} 
Notice how this procedure may return empty sets for prediction vectors close to decision boundaries, a behavior considered undesirable in the literature. 

\paragraph{Adaptive Prediction Sets} Subsequently, the Adaptive Prediction Sets (APS) method was introduced in \cite{romano2020classification}. It uses a different conformal score, which is the cumulated sum of the components of the soft label vector $y_i$, sorted in decreasing order, up to the component corresponding to the ground-truth label. Using the same notations as for TopK, let $q^{*}$ the rank of the ground truth labels in the sorted prediction vector, i.e. the integer such that $P_i^{-1}(q^{*}) = k^{*}$. The APS score then reads 
\begin{equation}
s^{\text{APS}} (k^{*}_i) = \sum_{q = 1}^{q^{*}} y_{i, P_i^{-1}(q)}.
\end{equation}
This procedure improves on the problem of non-empty conformal sets, but comes at the expense of prediction sets much larger than those constructed by the LAC method. 
\paragraph{Regularized Adaptive Prediction Sets} As a result, regularized APS (RAPS) was introduced in \cite{angelopoulos2020uncertainty}. It introduces two additional parameters, $k_{\text{reg}} \in \mathbb{N}$ and $\lambda_{\text{reg}} \in \mathbb{R}$, which penalize excessive set sizes. In detail, the APS score is modified as 
\begin{equation}
s_i^{\text{RAPS}}(k^{*}) =  s_i^{\text{APS}} + \lambda_{\text{reg}}\sum_{q=1}^{q^{*}} \mathbbm{1}_{q > k_{\text{reg}}}.
\end{equation}
The parameters $k_{\text{reg}}$ and $\lambda_{\text{reg}}$ are estimated from the data  by splitting the calibration set into two additional subsets. One is dedicated to the computation of these parameters, the other to the computation of the quantile. We follow the same setup as \cite{angelopoulos2020uncertainty} in our implementation.

\subsection{Zero-Shot Prediction with Vision-Language Models}
 Generally, a VLM projects both images and textual descriptions into a common latent space, enabling measurement of their similarities. The textual descriptions of target classes provided by the user, so called textual prompts, are transformed into numerical tokens ${\mathbf c}_k$. The latter are then mapped by the textual encoder to normalized embeddings ${\mathbf t}_k$ on the unit-hypersphere of $\mathbb{R}^d$, where $d$ is the latent space dimension. Similarly, the image $\mathbf{x}_i$ is processed by the visual encoder to produce embeddings $\boldsymbol{f}_i$ on the same unit-hypersphere. Prediction vectors are then obtained with a softmax transformation: 
 \begin{equation}
\label{zero-shot-prediction}
y_{i,k}= \frac{\exp ({\mathbf f}_i^T {\mathbf t}_k/T)}{\sum_{j}^K \exp ({\mathbf f}_i^T {\mathbf t}_k/T)}  
\end{equation} where $T$ is a parameter of the model, typically set to 0.01. The \textit{Zero-Shot} prediction is then $\hat{k} = \mathrm{argmax}_{k}~y_{i,k}$. In this paper, we use models trained within the CLIP framework \cite{radford_learning_2021}. We investigate four different models, characterized by their visual encoder, either be ViT~\cite{dosovitskiy2020image} based  (ViT-B/16, ViT-L/14) or CNN~\cite{lecun1995convolutional} based (RN50, RN101). We use the prompts "a photo of a \{class\}." for all experiments. Notice how $\boldsymbol{f}_i^{T} \boldsymbol{f}_j$ is an intuitive choice to measure the similarity between images $\boldsymbol{x}_i$ and $\boldsymbol{x}_j$. The latter has been used successfully in the few-shot adaptation literature, for instance by Tip-Adapter~\cite{zhang2021tip}.
\subsection{Localized conformal prediction}
\label{sec:localized}
We implement the algorithm presented in \cite{guan2023localized}.
For a given test sample \( \mathbf x_{\text{test}} \) and its corresponding projection in a latent space $\boldsymbol{f}_{\text{test}}$, LCP assigns greater importance to calibration examples that are more similar to it, using a localizer function \( H(\boldsymbol{f}_{\text{test}}, \boldsymbol{f}_i) \in [0, 1] \). This yields a test-specific, weighted empirical distribution over conformal scores, resulting in prediction sets that better adapt to local structure in the feature space.

In particular, given a conformal score $s_i$, $i =1,\dots, n$ for each calibration example, the weighted empirical distribution of scores is given by:

\begin{equation}
    \hat{\mathcal{F}}
    (\boldsymbol{f}_{\text{test}}) = \sum_{i=1}^n \frac{H(\boldsymbol{f}_{\text{test}}, \boldsymbol{f}_i)}{\sum_{j=1}^n H(\boldsymbol{f}_{\text{test}}, \boldsymbol{f}_j)} \, \delta_{s_i},
\end{equation}

where \( \delta_{s_i} \) is a point mass at the score \( s_i \).

For classification with label space \( \mathcal{Y} = \{1, \dots, K\} \), a class \( k \) is included in the prediction set for the test point if its test-time score \( s_{\text{test}}(k) \) does not exceed a quantile threshold:

\begin{equation}
    C(x_{\text{test}}) = \left\{ k \in \mathcal{Y} : s_{\text{test}}(k) \leq q_{\tilde{\alpha}} \right\},
\end{equation}

where $q_{\tilde{\alpha}}$ is the \( 1-\tilde{\alpha} \) quantile of the weighted distribution, and \( \tilde{\alpha} \) is chosen to ensure a marginal coverage of $1-\alpha$. We refer to \cite{guan2023localized} for additional details regarding the algorithm and technical proofs about the coverage guarantee of LCP.


\section{Methods}\label{sec:methods}

\subsection{Experimental setting}
\paragraph{Datasets} We follow the settings of previous works~\cite{zhou2022learning} and use 9 diverse images classification datasets: SUN397~\cite{sun397} for classification of scenes,  Aircraft~\cite{aircraft} for aircrafts, EuroSAT~\cite{eurosat} for satellite images, StanfordCars~\cite{cars} for cars models, Food101~\cite{food} for food items, Pets~\cite{pets} for pet breeds, Flower102~\cite{flower} for flowers species, DTD~\cite{dtd} for textures and UCF101~\cite{ucf101} for actions recognition. We exclude Caltech101~\cite{caltech101} from the benchmark, as most backbones achieves an accuracy higher than the target coverage $1 - \alpha$ for $\alpha = 0.1$. Details about the number of samples and labels distribution is provided in Supplementary Section \ref{sec:supp_datasets}.

\paragraph{Calibration sets generation} We uniformly sample 1000 images from the dataset for calibration, and use the remaining data to measure the metrics exposed in the next Section. Therefore, the calibration have the same label distribution as the dataset (see Supplementary Section \ref{sec:supp_datasets}). We use ten random seeds for each dataset, and use the same calibration / test splits for all methods. 

\paragraph{Metrics}
The CP literature usually wants to balance two objectives: (1) predictive efficiency (\textit{i.e.}, producing small prediction sets); and (2) conditional validity (\textit{i.e.}, maintaining the coverage guarantee across data subgroups).

Efficiency is measured by the \textit{mean set size} over the set of test samples $\mathcal{S}_{\text{test}}$:
\begin{equation}
    \text{Mean Set Size}(\mathcal{S}_{\text{test}}) = \frac{1}{|\mathcal{S}_{\text{test}}|} \sum_{\mathbf x \in \mathcal{S}_{\text{test}}} |C(\mathbf x)|.
\label{eq:mean_set_size}
\end{equation}

To assess conditional validity, we partition the test set by true class labels: $\mathcal{P} = \{\mathcal{S}_{\text{group}}\}$. For each group, the coverage Cov is:
\begin{equation}
    \text{Cov}(\mathcal{S}_{\text{group}}) = \frac{1}{|\mathcal{S}_{\text{group}}|} \sum_{(\mathbf x, y) \in \mathcal{S}_{\text{group}}} \mathbbm{1}_{y \in C(\mathbf x)}.
\end{equation}

We follow previous works \cite{fillioux2024foundation} and use two derived metrics:
\begin{itemize}
    \item \textbf{CovGap:} average deviation from the target coverage:
    \begin{equation}
        \text{CovGap}(\mathcal{S}_{\text{test}}) = \frac{1}{|\mathcal{P}|} \sum_{\mathcal{S}_{\text{group}} \in \mathcal{P}} \left| \text{Cov}(\mathcal{S}_{\text{group}}) - (1 - \alpha) \right|.
    \label{eq:covgap}
    \end{equation}
    
    \item \textbf{MCCC:} worst-case coverage across groups:
    \begin{equation}
        \text{MCCC}(\mathcal{S}_{\text{test}}) = \min_{\mathcal{S}_{\text{group}} \in \mathcal{P}} \text{Cov}(\mathcal{S}_{\text{group}}).
    \label{eq:MCCC}
    \end{equation}
\end{itemize}


\subsection{Transformation of Cosine Similarity}
\label{sec:crossval_sigmoid}
To define the localizer $H$ in the LCP framework, we apply a normalized sigmoid transformation to the cosine similarity between samples:
\begin{equation}
    H(\boldsymbol{f}_i, \boldsymbol{f}_j) = \frac{1 + \exp(-m(1 - \tau))}{1 + \exp(-m(\boldsymbol{f}_i^T \boldsymbol{f}_j - \tau))},
    \label{eq:sigmoid}
\end{equation}
where $m > 0$ controls the sigmoid's steepness, and $\tau \in [0, 1]$ sets its inflection point. 
We tune $m$ and $\tau$ via 5-fold cross-validation on the calibration set, so as to minimize the average set size. Finally, we take the median of the selected $m$ values and the mean of the selected $\tau$ values across folds. In our experiments, we search over $\tau \in \{0.8, 0.9, 1.0\}$ and $m \in \{1, 2, \ldots, 30\}$.

\section{Results}\label{sec:results}
\begin{figure}[t]
\begin{center}
\includegraphics[width=0.5\textwidth]{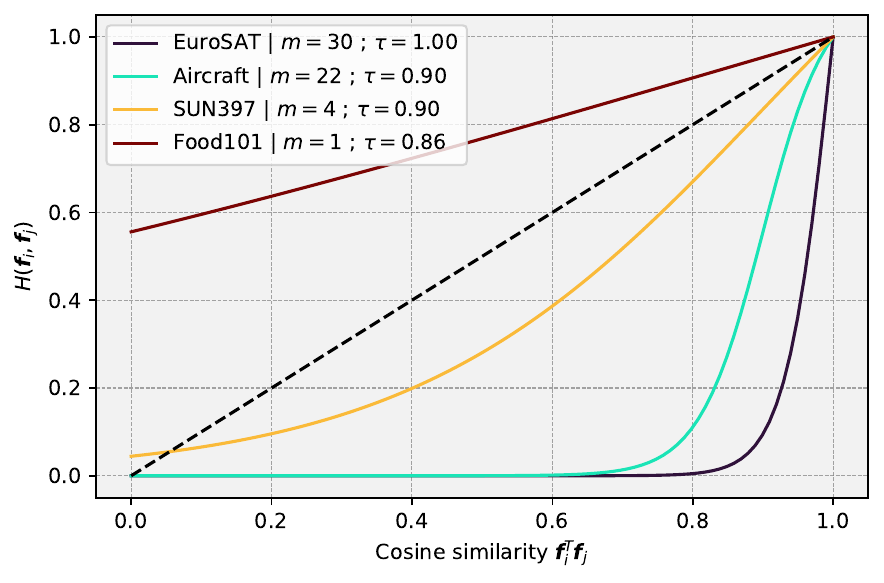}
\end{center}
\vspace{-10pt}
\caption{Examples of optimal sigmoid transformations (see Eq.~\ref{eq:sigmoid}) obtained with our cross validation procedure (see Section~\ref{sec:crossval_sigmoid}). The dotted black line corresponds to $H(\boldsymbol{f}_i, \boldsymbol{f}_j) = \boldsymbol{f}_i^T \boldsymbol{f}_j$, the naive version of the localized algorithm.}\label{fig:mtaus}
\vspace{-12pt}
\end{figure}
\subsection{Mean set sizes}
We present results concerning the mean set sizes in Table~\ref{tab:results_mean}. To assess the significance of the differences in mean set sizes between local methods and non-local baselines, we perform a paired t-test \cite{student1908probable} on the per-fold difference when it succeeds a Shapiro gaussianity test \cite{shapiro1965analysis}, and a Wilcoxon test \cite{wilcoxon1992individual} otherwise. We see that our approach achieves a decrease in mean set sizes for 9 out of 9 datasets for 3 out of 4 backbones compared to the non-local TopK, with statistical significance levels of least 0.01. Conversely, the naive approach fails to decrease the mean set sizes significantly compared to the non-local TopK, and even increases their sizes with statistical significance on several datasets and backbones (\textit{e.g.} DTD with ViT-B/16, or UCF101 with RN50). This trend is confirmed with other non-local baselines, although not as sharply as with TopK. We see that the naive approach degrades mean set sizes with statistical significance for several other conformal scores, such as APS for Aircraft with ViT-B/16. However, our approach achieves a statistically significant decrease of the mean set size in the latter setting, from 18.11 for the non-local baseline down to 17.43.

\aboverulesep = 0.1mm 
\belowrulesep = 0.1mm 
\begin{table*}
    \centering
    \caption{Averaged set sizes (see Eq. \ref{eq:mean_set_size}) over 10 folds, for $\alpha = 0.1$. The best (\textit{i.e}., lowest) values are highlighted in \textbf{bold}. Statistical significance in the performance difference at the $0.05$, $0.01$ and $0.001$ levels for a paired t-test (resp. Wilcoxon test) w.r.t. the non-local baseline are denoted by $^*$ (resp.~$^{\circ}$), $^{**}$ (resp.~$^{\circ\circ}$), and $^{***}$ (resp. $^{\circ\circ\circ}$). L- denotes local approaches (see Section \ref{sec:localized}). Naive corresponds to the black dotted line in Fig. \ref{fig:mtaus}, while the methods highlighted in \textcolor{Salmon}{\textbf{pink}} corresponds to our procedure described in Section \ref{sec:crossval_sigmoid}. Standard deviation is noted after the $\pm$ signs.}
    \vspace{-5pt}
\label{tab:results_mean}

\begin{subtable}{\linewidth}
        \setlength\dashlinedash{0.2pt}
                            \setlength\dashlinegap{1.5pt}
                            \setlength\arrayrulewidth{0.3pt}
                            \renewcommand{\arraystretch}{1.}
                                \caption{
            Results with ViT-B/16.}
        
        \centering
        \resizebox{\textwidth}{!}{
            \begin{tabular}{l|ccccccccc}
              \toprule

 & \rotatebox[origin=c]{45}{UCF101} & \rotatebox[origin=c]{45}{DTD} & \rotatebox[origin=c]{45}{Pets} & \rotatebox[origin=c]{45}{EuroSAT} & \rotatebox[origin=c]{45}{StanfordCars} & \rotatebox[origin=c]{45}{Flower102} & \rotatebox[origin=c]{45}{Aircraft} & \rotatebox[origin=c]{45}{SUN397} & \rotatebox[origin=c]{45}{Food101} 

\\ \midrule

\textbf{Zero-Shot Accuracy} & $\text{67.5}$ & $\text{43.3}$ & $\text{89.1}$ & $\text{48.3}$ & $\text{65.6}$ & $\text{70.8}$ & $\text{24.9}$ & $\text{62.6}$ & $\text{85.9}$ \\ \midrule

 TOPK &\text{\textcolor{black}{5.90}}\fontsize{8}{8}{\thinspace$\pm 0.74$}&\text{\textcolor{black}{19.00}}\fontsize{8}{8}{\thinspace$\pm 1.33$}&\text{\textcolor{black}{2.00}}\fontsize{8}{8}{\thinspace$\pm 0.00$}&\text{\textcolor{black}{6.00}}\fontsize{8}{8}{\thinspace$\pm 0.00$}&\text{\textcolor{black}{3.80}}\fontsize{8}{8}{\thinspace$\pm 0.42$}&\text{\textcolor{black}{9.50}}\fontsize{8}{8}{\thinspace$\pm 0.85$}&\text{\textcolor{black}{22.00}}\fontsize{8}{8}{\thinspace$\pm 0.82$}&\text{\textcolor{black}{4.80}}\fontsize{8}{8}{\thinspace$\pm 0.42$}&\text{\textcolor{black}{2.00}}\fontsize{8}{8}{\thinspace$\pm 0.00$}

\\

 L-TOPK (naive) &\text{\textcolor{black}{6.23}}$^{\circ}$\fontsize{8}{8}{\thinspace$\pm 0.63$}&\text{\textcolor{black}{19.98}}$^{\circ\circ\circ}$\fontsize{8}{8}{\thinspace$\pm 1.10$}&\text{\textcolor{black}{1.19}}$^{\circ\circ\circ}$\fontsize{8}{8}{\thinspace$\pm 0.21$}&\text{\textcolor{black}{5.31}}$^{\circ\circ\circ}$\fontsize{8}{8}{\thinspace$\pm 0.19$}&\text{\textcolor{black}{3.80}}\fontsize{8}{8}{\thinspace$\pm 0.42$}&\text{\textcolor{black}{8.27}}$^{\circ\circ\circ}$\fontsize{8}{8}{\thinspace$\pm 0.80$}&\text{\textcolor{black}{22.36}}$^{\circ}$\fontsize{8}{8}{\thinspace$\pm 0.82$}&\text{\textcolor{black}{4.50}}$^{\circ}$\fontsize{8}{8}{\thinspace$\pm 0.44$}&\text{\textcolor{black}{2.00}}\fontsize{8}{8}{\thinspace$\pm 0.00$}

\\

 \cellcolor{pink}L-TOPK &\cellcolor{pink} \fontsize{10}{11}{\textbf{\textcolor{black}{4.72}}}$^{\circ\circ\circ}$\fontsize{8}{8}{\thinspace$\pm 0.40$}&\cellcolor{pink} \fontsize{10}{11}{\textbf{\textcolor{black}{16.71}}}$^{\circ\circ\circ}$\fontsize{8}{8}{\thinspace$\pm 0.87$}&\cellcolor{pink} \fontsize{10}{11}{\textbf{\textcolor{black}{1.17}}}$^{\circ\circ\circ}$\fontsize{8}{8}{\thinspace$\pm 0.06$}&\cellcolor{pink} \fontsize{10}{11}{\textbf{\textcolor{black}{4.28}}}$^{\circ\circ\circ}$\fontsize{8}{8}{\thinspace$\pm 0.13$}&\cellcolor{pink} \fontsize{10}{11}{\textbf{\textcolor{black}{3.26}}}$^{\circ\circ\circ}$\fontsize{8}{8}{\thinspace$\pm 0.25$}&\cellcolor{pink} \fontsize{10}{11}{\textbf{\textcolor{black}{7.57}}}$^{\circ\circ\circ}$\fontsize{8}{8}{\thinspace$\pm 0.62$}&\cellcolor{pink} \fontsize{10}{11}{\textbf{\textcolor{black}{20.10}}}$^{\circ\circ\circ}$\fontsize{8}{8}{\thinspace$\pm 0.80$}&\cellcolor{pink} \fontsize{10}{11}{\textbf{\textcolor{black}{4.38}}}$^{\circ\circ}$\fontsize{8}{8}{\thinspace$\pm 0.37$}&\cellcolor{pink} \fontsize{10}{11}{\textbf{\textcolor{black}{1.63}}}$^{\circ\circ\circ}$\fontsize{8}{8}{\thinspace$\pm 0.17$}

\\

\midrule

 LAC &\text{\textcolor{black}{3.42}}\fontsize{8}{8}{\thinspace$\pm 0.25$}&\text{\textcolor{black}{15.76}}\fontsize{8}{8}{\thinspace$\pm 1.24$}&\fontsize{10}{11}{\textbf{\textcolor{black}{1.06}}}\fontsize{8}{8}{\thinspace$\pm 0.03$}&\text{\textcolor{black}{4.20}}\fontsize{8}{8}{\thinspace$\pm 0.06$}&\fontsize{10}{11}{\textbf{\textcolor{black}{2.46}}}\fontsize{8}{8}{\thinspace$\pm 0.11$}&\text{\textcolor{black}{5.50}}\fontsize{8}{8}{\thinspace$\pm 0.66$}&\text{\textcolor{black}{18.11}}\fontsize{8}{8}{\thinspace$\pm 0.96$}&\text{\textcolor{black}{3.47}}\fontsize{8}{8}{\thinspace$\pm 0.39$}&\fontsize{10}{11}{\textbf{\textcolor{black}{1.20}}}\fontsize{8}{8}{\thinspace$\pm 0.05$}

\\

 L-LAC (naive) &\text{\textcolor{black}{3.80}}$^{\circ\circ\circ}$\fontsize{8}{8}{\thinspace$\pm 0.29$}&\text{\textcolor{black}{16.97}}$^{\circ\circ\circ}$\fontsize{8}{8}{\thinspace$\pm 1.04$}&\text{\textcolor{black}{1.07}}$^{\circ}$\fontsize{8}{8}{\thinspace$\pm 0.04$}&\text{\textcolor{black}{4.16}}$^{\circ\circ\circ}$\fontsize{8}{8}{\thinspace$\pm 0.07$}&\text{\textcolor{black}{2.63}}$^{\circ\circ\circ}$\fontsize{8}{8}{\thinspace$\pm 0.16$}&\text{\textcolor{black}{4.93}}$^{\circ\circ\circ}$\fontsize{8}{8}{\thinspace$\pm 0.52$}&\text{\textcolor{black}{19.07}}$^{\circ\circ\circ}$\fontsize{8}{8}{\thinspace$\pm 0.87$}&\text{\textcolor{black}{3.54}}$^{\circ\circ}$\fontsize{8}{8}{\thinspace$\pm 0.37$}&\text{\textcolor{black}{1.29}}$^{\circ\circ\circ}$\fontsize{8}{8}{\thinspace$\pm 0.07$}

\\

 \cellcolor{pink}L-LAC &\cellcolor{pink} \fontsize{10}{11}{\textbf{\textcolor{black}{3.32}}}$^{\circ\circ}$\fontsize{8}{8}{\thinspace$\pm 0.27$}&\cellcolor{pink} \fontsize{10}{11}{\textbf{\textcolor{black}{14.62}}}$^{\circ\circ\circ}$\fontsize{8}{8}{\thinspace$\pm 0.93$}&\cellcolor{pink} \text{\textcolor{black}{1.09}}$^{\circ}$\fontsize{8}{8}{\thinspace$\pm 0.05$}&\cellcolor{pink} \fontsize{10}{11}{\textbf{\textcolor{black}{3.94}}}$^{\circ\circ\circ}$\fontsize{8}{8}{\thinspace$\pm 0.13$}&\cellcolor{pink} \text{\textcolor{black}{2.46}}\fontsize{8}{8}{\thinspace$\pm 0.17$}&\cellcolor{pink} \fontsize{10}{11}{\textbf{\textcolor{black}{4.92}}}$^{\circ\circ\circ}$\fontsize{8}{8}{\thinspace$\pm 0.44$}&\cellcolor{pink} \fontsize{10}{11}{\textbf{\textcolor{black}{17.43}}}$^{\circ}$\fontsize{8}{8}{\thinspace$\pm 0.81$}&\cellcolor{pink} \fontsize{10}{11}{\textbf{\textcolor{black}{3.44}}}\fontsize{8}{8}{\thinspace$\pm 0.35$}&\cellcolor{pink} \text{\textcolor{black}{1.41}}\fontsize{8}{8}{\thinspace$\pm 0.62$}

\\

\midrule

 APS &\text{\textcolor{black}{6.41}}\fontsize{8}{8}{\thinspace$\pm 0.34$}&\text{\textcolor{black}{16.64}}\fontsize{8}{8}{\thinspace$\pm 0.88$}&\text{\textcolor{black}{1.63}}\fontsize{8}{8}{\thinspace$\pm 0.03$}&\text{\textcolor{black}{4.63}}\fontsize{8}{8}{\thinspace$\pm 0.12$}&\text{\textcolor{black}{3.72}}\fontsize{8}{8}{\thinspace$\pm 0.18$}&\text{\textcolor{black}{6.41}}\fontsize{8}{8}{\thinspace$\pm 0.55$}&\text{\textcolor{black}{18.48}}\fontsize{8}{8}{\thinspace$\pm 0.62$}&\text{\textcolor{black}{7.68}}\fontsize{8}{8}{\thinspace$\pm 0.82$}&\fontsize{10}{11}{\textbf{\textcolor{black}{2.32}}}\fontsize{8}{8}{\thinspace$\pm 0.11$}

\\

 L-APS (naive) &\text{\textcolor{black}{6.80}}$^{\circ\circ\circ}$\fontsize{8}{8}{\thinspace$\pm 0.44$}&\text{\textcolor{black}{17.94}}$^{\circ\circ\circ}$\fontsize{8}{8}{\thinspace$\pm 0.77$}&\text{\textcolor{black}{1.65}}$^{\circ\circ}$\fontsize{8}{8}{\thinspace$\pm 0.03$}&\text{\textcolor{black}{4.58}}$^{\circ\circ\circ}$\fontsize{8}{8}{\thinspace$\pm 0.12$}&\text{\textcolor{black}{3.89}}$^{\circ\circ\circ}$\fontsize{8}{8}{\thinspace$\pm 0.18$}&\fontsize{10}{11}{\textbf{\textcolor{black}{5.85}}}$^{\circ\circ\circ}$\fontsize{8}{8}{\thinspace$\pm 0.44$}&\text{\textcolor{black}{19.31}}$^{\circ\circ\circ}$\fontsize{8}{8}{\thinspace$\pm 0.56$}&\text{\textcolor{black}{7.77}}$^{\circ\circ}$\fontsize{8}{8}{\thinspace$\pm 0.84$}&\text{\textcolor{black}{2.49}}$^{\circ\circ\circ}$\fontsize{8}{8}{\thinspace$\pm 0.12$}

\\

 \cellcolor{pink}L-APS &\cellcolor{pink} \fontsize{10}{11}{\textbf{\textcolor{black}{6.20}}}$^{\circ}$\fontsize{8}{8}{\thinspace$\pm 0.41$}&\cellcolor{pink} \fontsize{10}{11}{\textbf{\textcolor{black}{15.86}}}$^{\circ\circ\circ}$\fontsize{8}{8}{\thinspace$\pm 0.73$}&\cellcolor{pink} \fontsize{10}{11}{\textbf{\textcolor{black}{1.63}}}\fontsize{8}{8}{\thinspace$\pm 0.03$}&\cellcolor{pink} \fontsize{10}{11}{\textbf{\textcolor{black}{4.52}}}$^{\circ\circ\circ}$\fontsize{8}{8}{\thinspace$\pm 0.11$}&\cellcolor{pink} \fontsize{10}{11}{\textbf{\textcolor{black}{3.70}}}\fontsize{8}{8}{\thinspace$\pm 0.19$}&\cellcolor{pink} \text{\textcolor{black}{6.15}}$^{\circ}$\fontsize{8}{8}{\thinspace$\pm 0.49$}&\cellcolor{pink} \fontsize{10}{11}{\textbf{\textcolor{black}{18.11}}}$^{\circ}$\fontsize{8}{8}{\thinspace$\pm 0.52$}&\cellcolor{pink} \fontsize{10}{11}{\textbf{\textcolor{black}{7.67}}}\fontsize{8}{8}{\thinspace$\pm 0.82$}&\cellcolor{pink} \text{\textcolor{black}{2.35}}\fontsize{8}{8}{\thinspace$\pm 0.15$}

\\

\midrule

 RAPS &\text{\textcolor{black}{5.10}}\fontsize{8}{8}{\thinspace$\pm 1.01$}&\text{\textcolor{black}{15.47}}\fontsize{8}{8}{\thinspace$\pm 2.74$}&\text{\textcolor{black}{1.65}}\fontsize{8}{8}{\thinspace$\pm 0.04$}&\text{\textcolor{black}{4.52}}\fontsize{8}{8}{\thinspace$\pm 0.44$}&\text{\textcolor{black}{3.55}}\fontsize{8}{8}{\thinspace$\pm 0.33$}&\text{\textcolor{black}{5.88}}\fontsize{8}{8}{\thinspace$\pm 1.89$}&\text{\textcolor{black}{15.77}}\fontsize{8}{8}{\thinspace$\pm 2.01$}&\fontsize{10}{11}{\textbf{\textcolor{black}{4.49}}}\fontsize{8}{8}{\thinspace$\pm 0.45$}&\text{\textcolor{black}{1.97}}\fontsize{8}{8}{\thinspace$\pm 0.10$}

\\

 L-RAPS (naive) &\text{\textcolor{black}{5.36}}$^{**}$\fontsize{8}{8}{\thinspace$\pm 1.13$}&\text{\textcolor{black}{16.01}}$^{**}$\fontsize{8}{8}{\thinspace$\pm 2.60$}&\text{\textcolor{black}{1.67}}$^{***}$\fontsize{8}{8}{\thinspace$\pm 0.05$}&\text{\textcolor{black}{4.46}}\fontsize{8}{8}{\thinspace$\pm 0.42$}&\text{\textcolor{black}{3.67}}$^{\circ\circ}$\fontsize{8}{8}{\thinspace$\pm 0.42$}&\fontsize{10}{11}{\textbf{\textcolor{black}{5.39}}}\fontsize{8}{8}{\thinspace$\pm 1.19$}&\text{\textcolor{black}{16.36}}$^{**}$\fontsize{8}{8}{\thinspace$\pm 2.12$}&\text{\textcolor{black}{4.54}}\fontsize{8}{8}{\thinspace$\pm 0.40$}&\text{\textcolor{black}{2.07}}$^{***}$\fontsize{8}{8}{\thinspace$\pm 0.15$}

\\

 \cellcolor{pink}L-RAPS &\cellcolor{pink} \fontsize{10}{11}{\textbf{\textcolor{black}{4.85}}}\fontsize{8}{8}{\thinspace$\pm 0.57$}&\cellcolor{pink} \fontsize{10}{11}{\textbf{\textcolor{black}{14.67}}}$^{*}$\fontsize{8}{8}{\thinspace$\pm 2.32$}&\cellcolor{pink} \fontsize{10}{11}{\textbf{\textcolor{black}{1.65}}}\fontsize{8}{8}{\thinspace$\pm 0.04$}&\cellcolor{pink} \fontsize{10}{11}{\textbf{\textcolor{black}{4.07}}}$^{**}$\fontsize{8}{8}{\thinspace$\pm 0.43$}&\cellcolor{pink} \fontsize{10}{11}{\textbf{\textcolor{black}{3.53}}}\fontsize{8}{8}{\thinspace$\pm 0.35$}&\cellcolor{pink} \text{\textcolor{black}{5.59}}\fontsize{8}{8}{\thinspace$\pm 1.33$}&\cellcolor{pink} \fontsize{10}{11}{\textbf{\textcolor{black}{15.34}}}$^{*}$\fontsize{8}{8}{\thinspace$\pm 1.64$}&\cellcolor{pink} \text{\textcolor{black}{4.50}}$^{*}$\fontsize{8}{8}{\thinspace$\pm 0.45$}&\cellcolor{pink} \fontsize{10}{11}{\textbf{\textcolor{black}{1.96}}}\fontsize{8}{8}{\thinspace$\pm 0.11$}

\\

\midrule

\end{tabular}} \end{subtable}

\vspace{0.3cm}

\begin{subtable}{\linewidth}
        \setlength\dashlinedash{0.2pt}
                            \setlength\dashlinegap{1.5pt}
                            \setlength\arrayrulewidth{0.3pt}
                            \renewcommand{\arraystretch}{1.}
                                \caption{
            Results with ViT-L/14.}
        \centering
        \resizebox{\textwidth}{!}{
            \begin{tabular}{l|ccccccccc}
              \toprule

 & \rotatebox[origin=c]{45}{UCF101} & \rotatebox[origin=c]{45}{DTD} & \rotatebox[origin=c]{45}{Pets} & \rotatebox[origin=c]{45}{EuroSAT} & \rotatebox[origin=c]{45}{StanfordCars} & \rotatebox[origin=c]{45}{Flower102} & \rotatebox[origin=c]{45}{Aircraft} & \rotatebox[origin=c]{45}{SUN397} & \rotatebox[origin=c]{45}{Food101} 

\\ \midrule

\textbf{Zero-Shot Accuracy} & $\text{75.1}$ & $\text{53.4}$ & $\text{93.5}$ & $\text{60.3}$ & $\text{76.9}$ & $\text{79.6}$ & $\text{32.5}$ & $\text{67.7}$ & $\text{90.9}$ \\ \midrule

 TOPK &\text{\textcolor{black}{3.40}}\fontsize{8}{8}{\thinspace$\pm 0.52$}&\text{\textcolor{black}{12.70}}\fontsize{8}{8}{\thinspace$\pm 0.67$}&\fontsize{10}{11}{\textbf{\textcolor{black}{1.00}}}\fontsize{8}{8}{\thinspace$\pm 0.00$}&\text{\textcolor{black}{3.50}}\fontsize{8}{8}{\thinspace$\pm 0.53$}&\text{\textcolor{black}{2.50}}\fontsize{8}{8}{\thinspace$\pm 0.53$}&\text{\textcolor{black}{4.10}}\fontsize{8}{8}{\thinspace$\pm 0.57$}&\text{\textcolor{black}{9.40}}\fontsize{8}{8}{\thinspace$\pm 0.52$}&\text{\textcolor{black}{4.10}}\fontsize{8}{8}{\thinspace$\pm 0.32$}&\text{\textcolor{black}{1.30}}\fontsize{8}{8}{\thinspace$\pm 0.48$}

\\

 L-TOPK (naive) &\text{\textcolor{black}{3.20}}\fontsize{8}{8}{\thinspace$\pm 0.35$}&\text{\textcolor{black}{13.48}}$^{***}$\fontsize{8}{8}{\thinspace$\pm 0.83$}&\text{\textcolor{black}{1.00}}\fontsize{8}{8}{\thinspace$\pm 0.00$}&\text{\textcolor{black}{2.90}}$^{\circ\circ}$\fontsize{8}{8}{\thinspace$\pm 0.15$}&\text{\textcolor{black}{2.41}}\fontsize{8}{8}{\thinspace$\pm 0.44$}&\text{\textcolor{black}{3.70}}$^{\circ}$\fontsize{8}{8}{\thinspace$\pm 0.47$}&\text{\textcolor{black}{9.56}}\fontsize{8}{8}{\thinspace$\pm 0.49$}&\text{\textcolor{black}{3.95}}\fontsize{8}{8}{\thinspace$\pm 0.09$}&\text{\textcolor{black}{1.30}}\fontsize{8}{8}{\thinspace$\pm 0.48$}

\\

 \cellcolor{pink}L-TOPK &\cellcolor{pink} \fontsize{10}{11}{\textbf{\textcolor{black}{2.63}}}$^{***}$\fontsize{8}{8}{\thinspace$\pm 0.23$}&\cellcolor{pink} \fontsize{10}{11}{\textbf{\textcolor{black}{11.68}}}$^{***}$\fontsize{8}{8}{\thinspace$\pm 0.52$}&\cellcolor{pink} \text{\textcolor{black}{1.00}}\fontsize{8}{8}{\thinspace$\pm 0.00$}&\cellcolor{pink} \fontsize{10}{11}{\textbf{\textcolor{black}{2.79}}}$^{\circ\circ\circ}$\fontsize{8}{8}{\thinspace$\pm 0.15$}&\cellcolor{pink} \fontsize{10}{11}{\textbf{\textcolor{black}{1.99}}}$^{\circ\circ}$\fontsize{8}{8}{\thinspace$\pm 0.17$}&\cellcolor{pink} \fontsize{10}{11}{\textbf{\textcolor{black}{2.78}}}$^{***}$\fontsize{8}{8}{\thinspace$\pm 0.19$}&\cellcolor{pink} \fontsize{10}{11}{\textbf{\textcolor{black}{8.40}}}$^{***}$\fontsize{8}{8}{\thinspace$\pm 0.36$}&\cellcolor{pink} \fontsize{10}{11}{\textbf{\textcolor{black}{3.59}}}$^{***}$\fontsize{8}{8}{\thinspace$\pm 0.26$}&\cellcolor{pink} \fontsize{10}{11}{\textbf{\textcolor{black}{1.06}}}\fontsize{8}{8}{\thinspace$\pm 0.10$}

\\

\midrule

 LAC &\fontsize{10}{11}{\textbf{\textcolor{black}{1.94}}}\fontsize{8}{8}{\thinspace$\pm 0.09$}&\text{\textcolor{black}{9.91}}\fontsize{8}{8}{\thinspace$\pm 0.60$}&\fontsize{10}{11}{\textbf{\textcolor{black}{0.95}}}\fontsize{8}{8}{\thinspace$\pm 0.02$}&\text{\textcolor{black}{2.46}}\fontsize{8}{8}{\thinspace$\pm 0.19$}&\fontsize{10}{11}{\textbf{\textcolor{black}{1.57}}}\fontsize{8}{8}{\thinspace$\pm 0.08$}&\text{\textcolor{black}{2.03}}\fontsize{8}{8}{\thinspace$\pm 0.22$}&\text{\textcolor{black}{7.40}}\fontsize{8}{8}{\thinspace$\pm 0.31$}&\text{\textcolor{black}{2.92}}\fontsize{8}{8}{\thinspace$\pm 0.27$}&\fontsize{10}{11}{\textbf{\textcolor{black}{0.98}}}\fontsize{8}{8}{\thinspace$\pm 0.03$}

\\

 L-LAC (naive) &\text{\textcolor{black}{2.06}}$^{***}$\fontsize{8}{8}{\thinspace$\pm 0.06$}&\text{\textcolor{black}{10.89}}$^{***}$\fontsize{8}{8}{\thinspace$\pm 0.72$}&\text{\textcolor{black}{0.96}}$^{***}$\fontsize{8}{8}{\thinspace$\pm 0.02$}&\fontsize{10}{11}{\textbf{\textcolor{black}{2.40}}}$^{***}$\fontsize{8}{8}{\thinspace$\pm 0.18$}&\text{\textcolor{black}{1.64}}$^{***}$\fontsize{8}{8}{\thinspace$\pm 0.09$}&\text{\textcolor{black}{2.05}}\fontsize{8}{8}{\thinspace$\pm 0.24$}&\text{\textcolor{black}{7.87}}$^{***}$\fontsize{8}{8}{\thinspace$\pm 0.40$}&\text{\textcolor{black}{2.97}}$^{**}$\fontsize{8}{8}{\thinspace$\pm 0.26$}&\text{\textcolor{black}{1.02}}$^{***}$\fontsize{8}{8}{\thinspace$\pm 0.03$}

\\

 \cellcolor{pink}L-LAC &\cellcolor{pink} \text{\textcolor{black}{1.94}}\fontsize{8}{8}{\thinspace$\pm 0.09$}&\cellcolor{pink} \fontsize{10}{11}{\textbf{\textcolor{black}{9.54}}}$^{**}$\fontsize{8}{8}{\thinspace$\pm 0.51$}&\cellcolor{pink} \text{\textcolor{black}{0.96}}\fontsize{8}{8}{\thinspace$\pm 0.03$}&\cellcolor{pink} \text{\textcolor{black}{2.41}}\fontsize{8}{8}{\thinspace$\pm 0.14$}&\cellcolor{pink} \text{\textcolor{black}{1.59}}\fontsize{8}{8}{\thinspace$\pm 0.10$}&\cellcolor{pink} \fontsize{10}{11}{\textbf{\textcolor{black}{1.97}}}\fontsize{8}{8}{\thinspace$\pm 0.15$}&\cellcolor{pink} \fontsize{10}{11}{\textbf{\textcolor{black}{7.24}}}\fontsize{8}{8}{\thinspace$\pm 0.37$}&\cellcolor{pink} \fontsize{10}{11}{\textbf{\textcolor{black}{2.89}}}$^{**}$\fontsize{8}{8}{\thinspace$\pm 0.27$}&\cellcolor{pink} \text{\textcolor{black}{1.05}}\fontsize{8}{8}{\thinspace$\pm 0.11$}

\\

\midrule

 APS &\text{\textcolor{black}{3.81}}\fontsize{8}{8}{\thinspace$\pm 0.19$}&\text{\textcolor{black}{11.09}}\fontsize{8}{8}{\thinspace$\pm 0.55$}&\text{\textcolor{black}{1.33}}\fontsize{8}{8}{\thinspace$\pm 0.01$}&\text{\textcolor{black}{3.23}}\fontsize{8}{8}{\thinspace$\pm 0.12$}&\fontsize{10}{11}{\textbf{\textcolor{black}{2.35}}}\fontsize{8}{8}{\thinspace$\pm 0.09$}&\fontsize{10}{11}{\textbf{\textcolor{black}{3.04}}}\fontsize{8}{8}{\thinspace$\pm 0.12$}&\text{\textcolor{black}{8.74}}\fontsize{8}{8}{\thinspace$\pm 0.34$}&\text{\textcolor{black}{7.27}}\fontsize{8}{8}{\thinspace$\pm 0.57$}&\fontsize{10}{11}{\textbf{\textcolor{black}{1.70}}}\fontsize{8}{8}{\thinspace$\pm 0.07$}

\\

 L-APS (naive) &\text{\textcolor{black}{3.90}}$^{***}$\fontsize{8}{8}{\thinspace$\pm 0.22$}&\text{\textcolor{black}{12.15}}$^{***}$\fontsize{8}{8}{\thinspace$\pm 0.73$}&\text{\textcolor{black}{1.35}}$^{***}$\fontsize{8}{8}{\thinspace$\pm 0.01$}&\text{\textcolor{black}{3.19}}$^{\circ\circ\circ}$\fontsize{8}{8}{\thinspace$\pm 0.12$}&\text{\textcolor{black}{2.44}}$^{***}$\fontsize{8}{8}{\thinspace$\pm 0.11$}&\text{\textcolor{black}{3.04}}\fontsize{8}{8}{\thinspace$\pm 0.12$}&\text{\textcolor{black}{9.22}}$^{***}$\fontsize{8}{8}{\thinspace$\pm 0.40$}&\text{\textcolor{black}{7.41}}\fontsize{8}{8}{\thinspace$\pm 0.58$}&\text{\textcolor{black}{1.81}}$^{***}$\fontsize{8}{8}{\thinspace$\pm 0.08$}

\\

 \cellcolor{pink}L-APS &\cellcolor{pink} \fontsize{10}{11}{\textbf{\textcolor{black}{3.71}}}$^{**}$\fontsize{8}{8}{\thinspace$\pm 0.20$}&\cellcolor{pink} \fontsize{10}{11}{\textbf{\textcolor{black}{10.86}}}$^{**}$\fontsize{8}{8}{\thinspace$\pm 0.48$}&\cellcolor{pink} \fontsize{10}{11}{\textbf{\textcolor{black}{1.33}}}$^{**}$\fontsize{8}{8}{\thinspace$\pm 0.01$}&\cellcolor{pink} \fontsize{10}{11}{\textbf{\textcolor{black}{3.11}}}$^{***}$\fontsize{8}{8}{\thinspace$\pm 0.08$}&\cellcolor{pink} \text{\textcolor{black}{2.35}}\fontsize{8}{8}{\thinspace$\pm 0.11$}&\cellcolor{pink} \text{\textcolor{black}{3.07}}\fontsize{8}{8}{\thinspace$\pm 0.10$}&\cellcolor{pink} \fontsize{10}{11}{\textbf{\textcolor{black}{8.57}}}$^{***}$\fontsize{8}{8}{\thinspace$\pm 0.36$}&\cellcolor{pink} \fontsize{10}{11}{\textbf{\textcolor{black}{7.21}}}\fontsize{8}{8}{\thinspace$\pm 0.48$}&\cellcolor{pink} \text{\textcolor{black}{1.74}}\fontsize{8}{8}{\thinspace$\pm 0.10$}

\\

\midrule

 RAPS &\text{\textcolor{black}{2.92}}\fontsize{8}{8}{\thinspace$\pm 0.43$}&\text{\textcolor{black}{9.41}}\fontsize{8}{8}{\thinspace$\pm 1.75$}&\text{\textcolor{black}{1.29}}\fontsize{8}{8}{\thinspace$\pm 0.02$}&\text{\textcolor{black}{3.23}}\fontsize{8}{8}{\thinspace$\pm 0.37$}&\fontsize{10}{11}{\textbf{\textcolor{black}{2.25}}}\fontsize{8}{8}{\thinspace$\pm 0.08$}&\text{\textcolor{black}{2.88}}\fontsize{8}{8}{\thinspace$\pm 0.28$}&\text{\textcolor{black}{7.87}}\fontsize{8}{8}{\thinspace$\pm 0.51$}&\text{\textcolor{black}{3.80}}\fontsize{8}{8}{\thinspace$\pm 0.12$}&\fontsize{10}{11}{\textbf{\textcolor{black}{1.44}}}\fontsize{8}{8}{\thinspace$\pm 0.06$}

\\

 L-RAPS (naive) &\text{\textcolor{black}{2.98}}$^{\circ\circ}$\fontsize{8}{8}{\thinspace$\pm 0.45$}&\text{\textcolor{black}{9.88}}$^{***}$\fontsize{8}{8}{\thinspace$\pm 1.82$}&\text{\textcolor{black}{1.31}}$^{***}$\fontsize{8}{8}{\thinspace$\pm 0.02$}&\text{\textcolor{black}{3.19}}$^{**}$\fontsize{8}{8}{\thinspace$\pm 0.35$}&\text{\textcolor{black}{2.32}}$^{***}$\fontsize{8}{8}{\thinspace$\pm 0.09$}&\text{\textcolor{black}{2.89}}\fontsize{8}{8}{\thinspace$\pm 0.26$}&\text{\textcolor{black}{8.14}}$^{***}$\fontsize{8}{8}{\thinspace$\pm 0.55$}&\text{\textcolor{black}{3.82}}$^{*}$\fontsize{8}{8}{\thinspace$\pm 0.11$}&\text{\textcolor{black}{1.50}}$^{\circ\circ\circ}$\fontsize{8}{8}{\thinspace$\pm 0.07$}

\\

 \cellcolor{pink}L-RAPS &\cellcolor{pink} \fontsize{10}{11}{\textbf{\textcolor{black}{2.88}}}\fontsize{8}{8}{\thinspace$\pm 0.38$}&\cellcolor{pink} \fontsize{10}{11}{\textbf{\textcolor{black}{9.22}}}$^{**}$\fontsize{8}{8}{\thinspace$\pm 1.70$}&\cellcolor{pink} \fontsize{10}{11}{\textbf{\textcolor{black}{1.29}}}$^{*}$\fontsize{8}{8}{\thinspace$\pm 0.02$}&\cellcolor{pink} \fontsize{10}{11}{\textbf{\textcolor{black}{3.04}}}$^{*}$\fontsize{8}{8}{\thinspace$\pm 0.39$}&\cellcolor{pink} \text{\textcolor{black}{2.26}}\fontsize{8}{8}{\thinspace$\pm 0.09$}&\cellcolor{pink} \fontsize{10}{11}{\textbf{\textcolor{black}{2.80}}}$^{*}$\fontsize{8}{8}{\thinspace$\pm 0.22$}&\cellcolor{pink} \fontsize{10}{11}{\textbf{\textcolor{black}{7.68}}}$^{*}$\fontsize{8}{8}{\thinspace$\pm 0.53$}&\cellcolor{pink} \fontsize{10}{11}{\textbf{\textcolor{black}{3.79}}}\fontsize{8}{8}{\thinspace$\pm 0.11$}&\cellcolor{pink} \text{\textcolor{black}{1.45}}\fontsize{8}{8}{\thinspace$\pm 0.07$}

\\

\midrule

\end{tabular}} \end{subtable}

\vspace{0.3cm}

\begin{subtable}{\linewidth}
        \setlength\dashlinedash{0.2pt}
                            \setlength\dashlinegap{1.5pt}
                            \setlength\arrayrulewidth{0.3pt}
                            \renewcommand{\arraystretch}{1.}
                                \caption{
            Results with RN50.}
        \centering
        \resizebox{\textwidth}{!}{
            \begin{tabular}{l|ccccccccc}
              \toprule

 & \rotatebox[origin=c]{45}{UCF101} & \rotatebox[origin=c]{45}{DTD} & \rotatebox[origin=c]{45}{Pets} & \rotatebox[origin=c]{45}{EuroSAT} & \rotatebox[origin=c]{45}{StanfordCars} & \rotatebox[origin=c]{45}{Flower102} & \rotatebox[origin=c]{45}{Aircraft} & \rotatebox[origin=c]{45}{SUN397} & \rotatebox[origin=c]{45}{Food101} 

\\ \midrule

\textbf{Zero-Shot Accuracy} & $\text{61.9}$ & $\text{42.8}$ & $\text{85.7}$ & $\text{36.1}$ & $\text{55.8}$ & $\text{66.0}$ & $\text{17.0}$ & $\text{58.8}$ & $\text{77.4}$ \\ \midrule

 TOPK &\text{\textcolor{black}{8.40}}\fontsize{8}{8}{\thinspace$\pm 0.52$}&\text{\textcolor{black}{19.90}}\fontsize{8}{8}{\thinspace$\pm 1.20$}&\text{\textcolor{black}{2.00}}\fontsize{8}{8}{\thinspace$\pm 0.00$}&\text{\textcolor{black}{5.60}}\fontsize{8}{8}{\thinspace$\pm 0.52$}&\text{\textcolor{black}{6.50}}\fontsize{8}{8}{\thinspace$\pm 0.53$}&\text{\textcolor{black}{12.50}}\fontsize{8}{8}{\thinspace$\pm 1.65$}&\text{\textcolor{black}{34.20}}\fontsize{8}{8}{\thinspace$\pm 1.93$}&\text{\textcolor{black}{6.80}}\fontsize{8}{8}{\thinspace$\pm 0.63$}&\text{\textcolor{black}{3.10}}\fontsize{8}{8}{\thinspace$\pm 0.32$}

\\

 L-TOPK (naive) &\text{\textcolor{black}{8.82}}$^{\circ\circ}$\fontsize{8}{8}{\thinspace$\pm 0.52$}&\text{\textcolor{black}{21.02}}$^{***}$\fontsize{8}{8}{\thinspace$\pm 1.43$}&\text{\textcolor{black}{2.00}}\fontsize{8}{8}{\thinspace$\pm 0.00$}&\text{\textcolor{black}{4.62}}$^{\circ\circ\circ}$\fontsize{8}{8}{\thinspace$\pm 0.49$}&\text{\textcolor{black}{6.49}}\fontsize{8}{8}{\thinspace$\pm 0.55$}&\text{\textcolor{black}{11.14}}$^{\circ\circ\circ}$\fontsize{8}{8}{\thinspace$\pm 1.69$}&\text{\textcolor{black}{34.53}}\fontsize{8}{8}{\thinspace$\pm 1.93$}&\fontsize{10}{11}{\textbf{\textcolor{black}{6.14}}}$^{***}$\fontsize{8}{8}{\thinspace$\pm 0.61$}&\text{\textcolor{black}{3.10}}\fontsize{8}{8}{\thinspace$\pm 0.32$}

\\

 \cellcolor{pink}L-TOPK &\cellcolor{pink} \fontsize{10}{11}{\textbf{\textcolor{black}{7.42}}}$^{\circ\circ\circ}$\fontsize{8}{8}{\thinspace$\pm 0.34$}&\cellcolor{pink} \fontsize{10}{11}{\textbf{\textcolor{black}{18.84}}}$^{***}$\fontsize{8}{8}{\thinspace$\pm 1.04$}&\cellcolor{pink} \fontsize{10}{11}{\textbf{\textcolor{black}{1.52}}}$^{***}$\fontsize{8}{8}{\thinspace$\pm 0.08$}&\cellcolor{pink} \fontsize{10}{11}{\textbf{\textcolor{black}{4.58}}}$^{\circ\circ\circ}$\fontsize{8}{8}{\thinspace$\pm 0.07$}&\cellcolor{pink} \fontsize{10}{11}{\textbf{\textcolor{black}{5.66}}}$^{***}$\fontsize{8}{8}{\thinspace$\pm 0.59$}&\cellcolor{pink} \fontsize{10}{11}{\textbf{\textcolor{black}{11.01}}}$^{***}$\fontsize{8}{8}{\thinspace$\pm 1.51$}&\cellcolor{pink} \fontsize{10}{11}{\textbf{\textcolor{black}{30.66}}}$^{***}$\fontsize{8}{8}{\thinspace$\pm 1.57$}&\cellcolor{pink} \text{\textcolor{black}{6.15}}$^{***}$\fontsize{8}{8}{\thinspace$\pm 0.63$}&\cellcolor{pink} \fontsize{10}{11}{\textbf{\textcolor{black}{2.73}}}$^{***}$\fontsize{8}{8}{\thinspace$\pm 0.24$}

\\

\midrule

 LAC &\text{\textcolor{black}{5.52}}\fontsize{8}{8}{\thinspace$\pm 0.35$}&\text{\textcolor{black}{17.52}}\fontsize{8}{8}{\thinspace$\pm 0.97$}&\fontsize{10}{11}{\textbf{\textcolor{black}{1.27}}}\fontsize{8}{8}{\thinspace$\pm 0.05$}&\text{\textcolor{black}{4.71}}\fontsize{8}{8}{\thinspace$\pm 0.17$}&\text{\textcolor{black}{4.50}}\fontsize{8}{8}{\thinspace$\pm 0.46$}&\text{\textcolor{black}{7.79}}\fontsize{8}{8}{\thinspace$\pm 0.74$}&\fontsize{10}{11}{\textbf{\textcolor{black}{28.15}}}\fontsize{8}{8}{\thinspace$\pm 1.57$}&\text{\textcolor{black}{4.82}}\fontsize{8}{8}{\thinspace$\pm 0.33$}&\fontsize{10}{11}{\textbf{\textcolor{black}{1.94}}}\fontsize{8}{8}{\thinspace$\pm 0.17$}

\\

 L-LAC (naive) &\text{\textcolor{black}{6.01}}$^{***}$\fontsize{8}{8}{\thinspace$\pm 0.39$}&\text{\textcolor{black}{18.63}}$^{***}$\fontsize{8}{8}{\thinspace$\pm 1.13$}&\text{\textcolor{black}{1.30}}$^{***}$\fontsize{8}{8}{\thinspace$\pm 0.05$}&\text{\textcolor{black}{4.56}}$^{***}$\fontsize{8}{8}{\thinspace$\pm 0.16$}&\text{\textcolor{black}{4.75}}$^{***}$\fontsize{8}{8}{\thinspace$\pm 0.44$}&\fontsize{10}{11}{\textbf{\textcolor{black}{7.34}}}$^{***}$\fontsize{8}{8}{\thinspace$\pm 0.85$}&\text{\textcolor{black}{28.82}}$^{***}$\fontsize{8}{8}{\thinspace$\pm 1.65$}&\text{\textcolor{black}{4.74}}$^{***}$\fontsize{8}{8}{\thinspace$\pm 0.36$}&\text{\textcolor{black}{2.19}}$^{***}$\fontsize{8}{8}{\thinspace$\pm 0.19$}

\\

 \cellcolor{pink}L-LAC &\cellcolor{pink} \fontsize{10}{11}{\textbf{\textcolor{black}{5.37}}}$^{**}$\fontsize{8}{8}{\thinspace$\pm 0.36$}&\cellcolor{pink} \fontsize{10}{11}{\textbf{\textcolor{black}{16.91}}}$^{***}$\fontsize{8}{8}{\thinspace$\pm 1.01$}&\cellcolor{pink} \text{\textcolor{black}{1.28}}\fontsize{8}{8}{\thinspace$\pm 0.07$}&\cellcolor{pink} \fontsize{10}{11}{\textbf{\textcolor{black}{4.40}}}$^{***}$\fontsize{8}{8}{\thinspace$\pm 0.08$}&\cellcolor{pink} \fontsize{10}{11}{\textbf{\textcolor{black}{4.49}}}\fontsize{8}{8}{\thinspace$\pm 0.53$}&\cellcolor{pink} \text{\textcolor{black}{7.60}}\fontsize{8}{8}{\thinspace$\pm 0.88$}&\cellcolor{pink} \text{\textcolor{black}{28.40}}\fontsize{8}{8}{\thinspace$\pm 1.67$}&\cellcolor{pink} \fontsize{10}{11}{\textbf{\textcolor{black}{4.73}}}$^{***}$\fontsize{8}{8}{\thinspace$\pm 0.36$}&\cellcolor{pink} \text{\textcolor{black}{2.01}}\fontsize{8}{8}{\thinspace$\pm 0.21$}

\\

\midrule

 APS &\text{\textcolor{black}{8.51}}\fontsize{8}{8}{\thinspace$\pm 0.53$}&\text{\textcolor{black}{18.37}}\fontsize{8}{8}{\thinspace$\pm 0.83$}&\text{\textcolor{black}{1.97}}\fontsize{8}{8}{\thinspace$\pm 0.05$}&\text{\textcolor{black}{5.49}}\fontsize{8}{8}{\thinspace$\pm 0.13$}&\fontsize{10}{11}{\textbf{\textcolor{black}{6.43}}}\fontsize{8}{8}{\thinspace$\pm 0.41$}&\text{\textcolor{black}{8.80}}\fontsize{8}{8}{\thinspace$\pm 0.70$}&\text{\textcolor{black}{29.74}}\fontsize{8}{8}{\thinspace$\pm 1.47$}&\text{\textcolor{black}{10.16}}\fontsize{8}{8}{\thinspace$\pm 0.98$}&\fontsize{10}{11}{\textbf{\textcolor{black}{3.76}}}\fontsize{8}{8}{\thinspace$\pm 0.31$}

\\

 L-APS (naive) &\text{\textcolor{black}{8.76}}$^{***}$\fontsize{8}{8}{\thinspace$\pm 0.62$}&\text{\textcolor{black}{19.33}}$^{***}$\fontsize{8}{8}{\thinspace$\pm 0.96$}&\text{\textcolor{black}{2.01}}$^{***}$\fontsize{8}{8}{\thinspace$\pm 0.05$}&\text{\textcolor{black}{5.33}}$^{***}$\fontsize{8}{8}{\thinspace$\pm 0.12$}&\text{\textcolor{black}{6.74}}$^{***}$\fontsize{8}{8}{\thinspace$\pm 0.47$}&\fontsize{10}{11}{\textbf{\textcolor{black}{8.56}}}$^{*}$\fontsize{8}{8}{\thinspace$\pm 0.61$}&\text{\textcolor{black}{30.75}}$^{***}$\fontsize{8}{8}{\thinspace$\pm 1.63$}&\fontsize{10}{11}{\textbf{\textcolor{black}{10.04}}}$^{*}$\fontsize{8}{8}{\thinspace$\pm 0.97$}&\text{\textcolor{black}{4.01}}$^{***}$\fontsize{8}{8}{\thinspace$\pm 0.31$}

\\

 \cellcolor{pink}L-APS &\cellcolor{pink} \fontsize{10}{11}{\textbf{\textcolor{black}{8.30}}}$^{**}$\fontsize{8}{8}{\thinspace$\pm 0.58$}&\cellcolor{pink} \fontsize{10}{11}{\textbf{\textcolor{black}{17.92}}}$^{**}$\fontsize{8}{8}{\thinspace$\pm 0.82$}&\cellcolor{pink} \fontsize{10}{11}{\textbf{\textcolor{black}{1.96}}}\fontsize{8}{8}{\thinspace$\pm 0.06$}&\cellcolor{pink} \fontsize{10}{11}{\textbf{\textcolor{black}{5.05}}}$^{***}$\fontsize{8}{8}{\thinspace$\pm 0.11$}&\cellcolor{pink} \text{\textcolor{black}{6.44}}\fontsize{8}{8}{\thinspace$\pm 0.41$}&\cellcolor{pink} \text{\textcolor{black}{8.94}}\fontsize{8}{8}{\thinspace$\pm 0.87$}&\cellcolor{pink} \fontsize{10}{11}{\textbf{\textcolor{black}{29.56}}}\fontsize{8}{8}{\thinspace$\pm 1.31$}&\cellcolor{pink} \text{\textcolor{black}{10.05}}$^{*}$\fontsize{8}{8}{\thinspace$\pm 0.98$}&\cellcolor{pink} \text{\textcolor{black}{3.76}}\fontsize{8}{8}{\thinspace$\pm 0.33$}

\\

\midrule

 RAPS &\text{\textcolor{black}{7.14}}\fontsize{8}{8}{\thinspace$\pm 1.11$}&\text{\textcolor{black}{16.22}}\fontsize{8}{8}{\thinspace$\pm 1.85$}&\text{\textcolor{black}{1.99}}\fontsize{8}{8}{\thinspace$\pm 0.05$}&\text{\textcolor{black}{5.04}}\fontsize{8}{8}{\thinspace$\pm 0.77$}&\text{\textcolor{black}{5.42}}\fontsize{8}{8}{\thinspace$\pm 0.62$}&\text{\textcolor{black}{6.97}}\fontsize{8}{8}{\thinspace$\pm 1.34$}&\fontsize{10}{11}{\textbf{\textcolor{black}{27.58}}}\fontsize{8}{8}{\thinspace$\pm 3.31$}&\text{\textcolor{black}{6.26}}\fontsize{8}{8}{\thinspace$\pm 0.72$}&\text{\textcolor{black}{3.09}}\fontsize{8}{8}{\thinspace$\pm 0.38$}

\\

 L-RAPS (naive) &\text{\textcolor{black}{7.38}}$^{**}$\fontsize{8}{8}{\thinspace$\pm 1.25$}&\text{\textcolor{black}{16.68}}$^{**}$\fontsize{8}{8}{\thinspace$\pm 1.76$}&\text{\textcolor{black}{2.01}}$^{\circ\circ}$\fontsize{8}{8}{\thinspace$\pm 0.05$}&\text{\textcolor{black}{4.94}}$^{***}$\fontsize{8}{8}{\thinspace$\pm 0.74$}&\text{\textcolor{black}{5.61}}$^{\circ\circ}$\fontsize{8}{8}{\thinspace$\pm 0.70$}&\fontsize{10}{11}{\textbf{\textcolor{black}{6.91}}}\fontsize{8}{8}{\thinspace$\pm 1.64$}&\text{\textcolor{black}{28.91}}\fontsize{8}{8}{\thinspace$\pm 3.66$}&\fontsize{10}{11}{\textbf{\textcolor{black}{6.24}}}\fontsize{8}{8}{\thinspace$\pm 0.74$}&\text{\textcolor{black}{3.34}}$^{\circ\circ}$\fontsize{8}{8}{\thinspace$\pm 0.52$}

\\

 \cellcolor{pink}L-RAPS &\cellcolor{pink} \fontsize{10}{11}{\textbf{\textcolor{black}{7.07}}}\fontsize{8}{8}{\thinspace$\pm 1.07$}&\cellcolor{pink} \fontsize{10}{11}{\textbf{\textcolor{black}{15.45}}}$^{**}$\fontsize{8}{8}{\thinspace$\pm 1.94$}&\cellcolor{pink} \fontsize{10}{11}{\textbf{\textcolor{black}{1.98}}}\fontsize{8}{8}{\thinspace$\pm 0.06$}&\cellcolor{pink} \fontsize{10}{11}{\textbf{\textcolor{black}{4.23}}}$^{***}$\fontsize{8}{8}{\thinspace$\pm 0.58$}&\cellcolor{pink} \fontsize{10}{11}{\textbf{\textcolor{black}{5.37}}}$^{*}$\fontsize{8}{8}{\thinspace$\pm 0.63$}&\cellcolor{pink} \text{\textcolor{black}{7.01}}\fontsize{8}{8}{\thinspace$\pm 1.32$}&\cellcolor{pink} \text{\textcolor{black}{27.72}}\fontsize{8}{8}{\thinspace$\pm 3.31$}&\cellcolor{pink} \text{\textcolor{black}{6.25}}\fontsize{8}{8}{\thinspace$\pm 0.76$}&\cellcolor{pink} \fontsize{10}{11}{\textbf{\textcolor{black}{3.08}}}\fontsize{8}{8}{\thinspace$\pm 0.39$}

\\

\midrule

\end{tabular}} \end{subtable}

\vspace{0.3cm}

\begin{subtable}{\linewidth}
        \setlength\dashlinedash{0.2pt}
                            \setlength\dashlinegap{1.5pt}
                            \setlength\arrayrulewidth{0.3pt}
                            \renewcommand{\arraystretch}{1.}
                                \caption{
            Results with RN101.}
        \centering
        \resizebox{\textwidth}{!}{
            \begin{tabular}{l|ccccccccc}
              \toprule

 & \rotatebox[origin=c]{45}{UCF101} & \rotatebox[origin=c]{45}{DTD} & \rotatebox[origin=c]{45}{Pets} & \rotatebox[origin=c]{45}{EuroSAT} & \rotatebox[origin=c]{45}{StanfordCars} & \rotatebox[origin=c]{45}{Flower102} & \rotatebox[origin=c]{45}{Aircraft} & \rotatebox[origin=c]{45}{SUN397} & \rotatebox[origin=c]{45}{Food101} 

\\ \midrule

\textbf{Zero-Shot Accuracy} & $\text{61.0}$ & $\text{37.1}$ & $\text{86.9}$ & $\text{32.8}$ & $\text{63.2}$ & $\text{64.4}$ & $\text{18.1}$ & $\text{59.0}$ & $\text{80.7}$ \\ \midrule

 TOPK &\text{\textcolor{black}{6.80}}\fontsize{8}{8}{\thinspace$\pm 0.63$}&\text{\textcolor{black}{19.00}}\fontsize{8}{8}{\thinspace$\pm 0.82$}&\text{\textcolor{black}{2.00}}\fontsize{8}{8}{\thinspace$\pm 0.00$}&\text{\textcolor{black}{8.10}}\fontsize{8}{8}{\thinspace$\pm 0.32$}&\text{\textcolor{black}{4.50}}\fontsize{8}{8}{\thinspace$\pm 0.53$}&\text{\textcolor{black}{10.10}}\fontsize{8}{8}{\thinspace$\pm 0.88$}&\text{\textcolor{black}{30.30}}\fontsize{8}{8}{\thinspace$\pm 1.49$}&\text{\textcolor{black}{6.70}}\fontsize{8}{8}{\thinspace$\pm 0.82$}&\text{\textcolor{black}{3.10}}\fontsize{8}{8}{\thinspace$\pm 0.32$}

\\

 L-TOPK (naive) &\text{\textcolor{black}{6.83}}\fontsize{8}{8}{\thinspace$\pm 0.66$}&\text{\textcolor{black}{18.47}}$^{\circ}$\fontsize{8}{8}{\thinspace$\pm 0.84$}&\fontsize{10}{11}{\textbf{\textcolor{black}{1.00}}}$^{***}$\fontsize{8}{8}{\thinspace$\pm 0.00$}&\text{\textcolor{black}{7.12}}$^{\circ\circ\circ}$\fontsize{8}{8}{\thinspace$\pm 0.31$}&\text{\textcolor{black}{4.39}}\fontsize{8}{8}{\thinspace$\pm 0.50$}&\fontsize{10}{11}{\textbf{\textcolor{black}{8.09}}}$^{***}$\fontsize{8}{8}{\thinspace$\pm 0.65$}&\fontsize{10}{11}{\textbf{\textcolor{black}{27.97}}}$^{***}$\fontsize{8}{8}{\thinspace$\pm 1.28$}&\text{\textcolor{black}{7.43}}$^{\circ\circ\circ}$\fontsize{8}{8}{\thinspace$\pm 0.59$}&\text{\textcolor{black}{2.76}}$^{\circ}$\fontsize{8}{8}{\thinspace$\pm 0.41$}

\\

 \cellcolor{pink}L-TOPK &\cellcolor{pink} \fontsize{10}{11}{\textbf{\textcolor{black}{5.73}}}$^{***}$\fontsize{8}{8}{\thinspace$\pm 0.33$}&\cellcolor{pink} \fontsize{10}{11}{\textbf{\textcolor{black}{17.84}}}$^{***}$\fontsize{8}{8}{\thinspace$\pm 0.90$}&\cellcolor{pink} \text{\textcolor{black}{1.50}}$^{***}$\fontsize{8}{8}{\thinspace$\pm 0.06$}&\cellcolor{pink} \fontsize{10}{11}{\textbf{\textcolor{black}{6.69}}}$^{***}$\fontsize{8}{8}{\thinspace$\pm 0.18$}&\cellcolor{pink} \fontsize{10}{11}{\textbf{\textcolor{black}{3.89}}}$^{***}$\fontsize{8}{8}{\thinspace$\pm 0.41$}&\cellcolor{pink} \text{\textcolor{black}{8.47}}$^{***}$\fontsize{8}{8}{\thinspace$\pm 0.77$}&\cellcolor{pink} \text{\textcolor{black}{28.90}}$^{***}$\fontsize{8}{8}{\thinspace$\pm 1.30$}&\cellcolor{pink} \fontsize{10}{11}{\textbf{\textcolor{black}{6.21}}}$^{**}$\fontsize{8}{8}{\thinspace$\pm 0.82$}&\cellcolor{pink} \fontsize{10}{11}{\textbf{\textcolor{black}{2.62}}}$^{***}$\fontsize{8}{8}{\thinspace$\pm 0.38$}

\\

\midrule

 LAC &\fontsize{10}{11}{\textbf{\textcolor{black}{4.50}}}\fontsize{8}{8}{\thinspace$\pm 0.39$}&\text{\textcolor{black}{17.05}}\fontsize{8}{8}{\thinspace$\pm 0.90$}&\text{\textcolor{black}{1.21}}\fontsize{8}{8}{\thinspace$\pm 0.03$}&\text{\textcolor{black}{7.08}}\fontsize{8}{8}{\thinspace$\pm 0.10$}&\fontsize{10}{11}{\textbf{\textcolor{black}{2.96}}}\fontsize{8}{8}{\thinspace$\pm 0.27$}&\text{\textcolor{black}{6.49}}\fontsize{8}{8}{\thinspace$\pm 0.87$}&\text{\textcolor{black}{27.37}}\fontsize{8}{8}{\thinspace$\pm 1.78$}&\text{\textcolor{black}{4.81}}\fontsize{8}{8}{\thinspace$\pm 0.56$}&\fontsize{10}{11}{\textbf{\textcolor{black}{1.76}}}\fontsize{8}{8}{\thinspace$\pm 0.16$}

\\

 L-LAC (naive) &\text{\textcolor{black}{4.82}}$^{\circ\circ\circ}$\fontsize{8}{8}{\thinspace$\pm 0.44$}&\text{\textcolor{black}{16.67}}$^{**}$\fontsize{8}{8}{\thinspace$\pm 1.00$}&\fontsize{10}{11}{\textbf{\textcolor{black}{1.17}}}$^{***}$\fontsize{8}{8}{\thinspace$\pm 0.03$}&\text{\textcolor{black}{6.98}}$^{***}$\fontsize{8}{8}{\thinspace$\pm 0.09$}&\text{\textcolor{black}{3.10}}$^{***}$\fontsize{8}{8}{\thinspace$\pm 0.25$}&\fontsize{10}{11}{\textbf{\textcolor{black}{5.45}}}$^{\circ\circ\circ}$\fontsize{8}{8}{\thinspace$\pm 0.72$}&\fontsize{10}{11}{\textbf{\textcolor{black}{25.80}}}$^{***}$\fontsize{8}{8}{\thinspace$\pm 1.70$}&\text{\textcolor{black}{5.41}}$^{***}$\fontsize{8}{8}{\thinspace$\pm 0.54$}&\text{\textcolor{black}{1.78}}\fontsize{8}{8}{\thinspace$\pm 0.17$}

\\

 \cellcolor{pink}L-LAC &\cellcolor{pink} \text{\textcolor{black}{4.55}}\fontsize{8}{8}{\thinspace$\pm 0.51$}&\cellcolor{pink} \fontsize{10}{11}{\textbf{\textcolor{black}{16.36}}}$^{***}$\fontsize{8}{8}{\thinspace$\pm 0.85$}&\cellcolor{pink} \text{\textcolor{black}{1.23}}$^{*}$\fontsize{8}{8}{\thinspace$\pm 0.05$}&\cellcolor{pink} \fontsize{10}{11}{\textbf{\textcolor{black}{6.64}}}$^{***}$\fontsize{8}{8}{\thinspace$\pm 0.09$}&\cellcolor{pink} \text{\textcolor{black}{3.06}}\fontsize{8}{8}{\thinspace$\pm 0.41$}&\cellcolor{pink} \text{\textcolor{black}{7.86}}$^{***}$\fontsize{8}{8}{\thinspace$\pm 0.65$}&\cellcolor{pink} \text{\textcolor{black}{26.55}}$^{**}$\fontsize{8}{8}{\thinspace$\pm 1.71$}&\cellcolor{pink} \fontsize{10}{11}{\textbf{\textcolor{black}{4.73}}}$^{*}$\fontsize{8}{8}{\thinspace$\pm 0.49$}&\cellcolor{pink} \text{\textcolor{black}{1.83}}$^{*}$\fontsize{8}{8}{\thinspace$\pm 0.19$}

\\

\midrule

 APS &\text{\textcolor{black}{8.07}}\fontsize{8}{8}{\thinspace$\pm 0.37$}&\text{\textcolor{black}{18.01}}\fontsize{8}{8}{\thinspace$\pm 0.79$}&\text{\textcolor{black}{1.86}}\fontsize{8}{8}{\thinspace$\pm 0.02$}&\text{\textcolor{black}{7.46}}\fontsize{8}{8}{\thinspace$\pm 0.19$}&\text{\textcolor{black}{4.38}}\fontsize{8}{8}{\thinspace$\pm 0.23$}&\text{\textcolor{black}{7.99}}\fontsize{8}{8}{\thinspace$\pm 0.70$}&\text{\textcolor{black}{28.97}}\fontsize{8}{8}{\thinspace$\pm 1.85$}&\text{\textcolor{black}{10.84}}\fontsize{8}{8}{\thinspace$\pm 0.91$}&\fontsize{10}{11}{\textbf{\textcolor{black}{3.36}}}\fontsize{8}{8}{\thinspace$\pm 0.15$}

\\

 L-APS (naive) &\text{\textcolor{black}{8.64}}$^{***}$\fontsize{8}{8}{\thinspace$\pm 0.39$}&\text{\textcolor{black}{17.57}}$^{**}$\fontsize{8}{8}{\thinspace$\pm 0.96$}&\fontsize{10}{11}{\textbf{\textcolor{black}{1.82}}}$^{***}$\fontsize{8}{8}{\thinspace$\pm 0.03$}&\text{\textcolor{black}{7.34}}$^{***}$\fontsize{8}{8}{\thinspace$\pm 0.17$}&\text{\textcolor{black}{4.61}}$^{***}$\fontsize{8}{8}{\thinspace$\pm 0.23$}&\fontsize{10}{11}{\textbf{\textcolor{black}{7.01}}}$^{***}$\fontsize{8}{8}{\thinspace$\pm 0.54$}&\fontsize{10}{11}{\textbf{\textcolor{black}{27.34}}}$^{***}$\fontsize{8}{8}{\thinspace$\pm 1.81$}&\text{\textcolor{black}{12.44}}$^{\circ\circ\circ}$\fontsize{8}{8}{\thinspace$\pm 0.91$}&\text{\textcolor{black}{3.45}}$^{*}$\fontsize{8}{8}{\thinspace$\pm 0.17$}

\\

 \cellcolor{pink}L-APS &\cellcolor{pink} \fontsize{10}{11}{\textbf{\textcolor{black}{7.75}}}$^{***}$\fontsize{8}{8}{\thinspace$\pm 0.35$}&\cellcolor{pink} \fontsize{10}{11}{\textbf{\textcolor{black}{17.41}}}$^{***}$\fontsize{8}{8}{\thinspace$\pm 0.67$}&\cellcolor{pink} \text{\textcolor{black}{1.87}}\fontsize{8}{8}{\thinspace$\pm 0.01$}&\cellcolor{pink} \fontsize{10}{11}{\textbf{\textcolor{black}{6.96}}}$^{***}$\fontsize{8}{8}{\thinspace$\pm 0.12$}&\cellcolor{pink} \fontsize{10}{11}{\textbf{\textcolor{black}{4.38}}}\fontsize{8}{8}{\thinspace$\pm 0.23$}&\cellcolor{pink} \text{\textcolor{black}{7.86}}\fontsize{8}{8}{\thinspace$\pm 0.65$}&\cellcolor{pink} \text{\textcolor{black}{28.24}}$^{**}$\fontsize{8}{8}{\thinspace$\pm 2.07$}&\cellcolor{pink} \fontsize{10}{11}{\textbf{\textcolor{black}{10.68}}}\fontsize{8}{8}{\thinspace$\pm 0.82$}&\cellcolor{pink} \text{\textcolor{black}{3.45}}\fontsize{8}{8}{\thinspace$\pm 0.29$}

\\

\midrule

 RAPS &\text{\textcolor{black}{5.39}}\fontsize{8}{8}{\thinspace$\pm 0.38$}&\text{\textcolor{black}{15.25}}\fontsize{8}{8}{\thinspace$\pm 2.13$}&\text{\textcolor{black}{1.85}}\fontsize{8}{8}{\thinspace$\pm 0.10$}&\text{\textcolor{black}{5.87}}\fontsize{8}{8}{\thinspace$\pm 0.89$}&\text{\textcolor{black}{3.95}}\fontsize{8}{8}{\thinspace$\pm 0.35$}&\text{\textcolor{black}{7.11}}\fontsize{8}{8}{\thinspace$\pm 2.27$}&\text{\textcolor{black}{25.58}}\fontsize{8}{8}{\thinspace$\pm 4.23$}&\text{\textcolor{black}{6.16}}\fontsize{8}{8}{\thinspace$\pm 0.92$}&\text{\textcolor{black}{2.81}}\fontsize{8}{8}{\thinspace$\pm 0.22$}

\\

 L-RAPS (naive) &\text{\textcolor{black}{5.64}}$^{***}$\fontsize{8}{8}{\thinspace$\pm 0.48$}&\text{\textcolor{black}{15.29}}\fontsize{8}{8}{\thinspace$\pm 1.97$}&\fontsize{10}{11}{\textbf{\textcolor{black}{1.81}}}$^{***}$\fontsize{8}{8}{\thinspace$\pm 0.09$}&\text{\textcolor{black}{5.71}}\fontsize{8}{8}{\thinspace$\pm 0.90$}&\text{\textcolor{black}{4.06}}$^{***}$\fontsize{8}{8}{\thinspace$\pm 0.37$}&\fontsize{10}{11}{\textbf{\textcolor{black}{6.21}}}$^{\circ}$\fontsize{8}{8}{\thinspace$\pm 1.15$}&\text{\textcolor{black}{24.88}}$^{**}$\fontsize{8}{8}{\thinspace$\pm 4.39$}&\text{\textcolor{black}{6.80}}$^{**}$\fontsize{8}{8}{\thinspace$\pm 1.28$}&\text{\textcolor{black}{2.83}}\fontsize{8}{8}{\thinspace$\pm 0.22$}

\\

 \cellcolor{pink}L-RAPS &\cellcolor{pink} \fontsize{10}{11}{\textbf{\textcolor{black}{5.38}}}\fontsize{8}{8}{\thinspace$\pm 0.41$}&\cellcolor{pink} \fontsize{10}{11}{\textbf{\textcolor{black}{15.17}}}\fontsize{8}{8}{\thinspace$\pm 1.94$}&\cellcolor{pink} \text{\textcolor{black}{1.83}}$^{\circ}$\fontsize{8}{8}{\thinspace$\pm 0.10$}&\cellcolor{pink} \fontsize{10}{11}{\textbf{\textcolor{black}{5.31}}}$^{***}$\fontsize{8}{8}{\thinspace$\pm 1.02$}&\cellcolor{pink} \fontsize{10}{11}{\textbf{\textcolor{black}{3.91}}}\fontsize{8}{8}{\thinspace$\pm 0.33$}&\cellcolor{pink} \text{\textcolor{black}{6.65}}\fontsize{8}{8}{\thinspace$\pm 1.46$}&\cellcolor{pink} \fontsize{10}{11}{\textbf{\textcolor{black}{24.85}}}$^{\circ}$\fontsize{8}{8}{\thinspace$\pm 3.52$}&\cellcolor{pink} \fontsize{10}{11}{\textbf{\textcolor{black}{6.14}}}\fontsize{8}{8}{\thinspace$\pm 0.88$}&\cellcolor{pink} \fontsize{10}{11}{\textbf{\textcolor{black}{2.79}}}\fontsize{8}{8}{\thinspace$\pm 0.21$}

\\

\midrule

\end{tabular}} \end{subtable}

\end{table*}

Overall, these results prove the effectiveness of our approach and the necessity of transforming the cosine similarities when using LCP. As an illustration, Figure~\ref{fig:mtaus} shows examples of sigmoidal transformations obtained with our cross validation strategy exposed in Section \ref{sec:crossval_sigmoid}. 
\subsection{Coverage metrics}
We present results concerning coverage metrics in Table~\ref{tab:results_cvg}, namely CovGap (see Eq.~\ref{eq:covgap}) and MCCC (see Eq. \ref{eq:MCCC}).
Compared to the mean set sizes, it is more difficult to identify trends that hold across datasets and backbones. As a general observation, the local algorithms seem to more closely follow the behavior of the non-local baselines. However, there are still several instances where our approach improves MCCC, \textit{e.g.}, for the LAC conformal score on Pets with ViT-B/16. However, on the same dataset with the same backbone, the MCCC is degraded compared to the non-local baseline when using the TopK conformal score. For the CovGap metric, the results obtained with the local algorithms are almost identical to the non-local baselines across all datasets and backbones. This result suggests that localization of conformal prediction is not sufficient to minimize the CovGap for image classification using VLMs.
\begin{table*}
    \centering
    \caption{CovGap / MCCC (see Eq. \ref{eq:covgap} and Eq. \ref{eq:MCCC}) averaged over 10 folds, for $\alpha = 0.1$. For CovGap, lower values are better, while for MCCC, higher values are better. The best results are highlighted in \textbf{bold}.  L- denotes local approaches (see Section \ref{sec:localized}). Naive corresponds to the black dotted line in Fig. \ref{fig:mtaus}, while the methods highlighted in \textcolor{Salmon}{\textbf{pink}} corresponds to our procedure described in Section \ref{sec:crossval_sigmoid}.}
    \vspace{-5pt}
\label{tab:results_cvg}

\begin{subtable}{\linewidth}
        \setlength\dashlinedash{0.2pt}
                            \setlength\dashlinegap{1.5pt}
                            \setlength\arrayrulewidth{0.3pt}
                            \renewcommand{\arraystretch}{0.85}
                                \caption{
            Results with ViT-B/16.}
        
        \centering
        \resizebox{\textwidth}{!}{
            \begin{tabular}{l|ccccccccc}
              \toprule

 & \rotatebox[origin=c]{45}{UCF101} & \rotatebox[origin=c]{45}{DTD} & \rotatebox[origin=c]{45}{Pets} & \rotatebox[origin=c]{45}{EuroSAT} & \rotatebox[origin=c]{45}{StanfordCars} & \rotatebox[origin=c]{45}{Flower102} & \rotatebox[origin=c]{45}{Aircraft} & \rotatebox[origin=c]{45}{SUN397} & \rotatebox[origin=c]{45}{Food101} 

\\ \midrule

\textbf{Zero-Shot Accuracy} & $\text{67.5}$ & $\text{43.3}$ & $\text{89.1}$ & $\text{48.3}$ & $\text{65.6}$ & $\text{70.8}$ & $\text{24.9}$ & $\text{62.6}$ & $\text{85.9}$ \\ \midrule

 TOPK &\text{\textcolor{black}{0.12}} / \text{\textcolor{black}{0.03}}&\text{\textcolor{black}{0.12}} / \text{\textcolor{black}{0.13}}&\text{\textcolor{black}{0.09}} / \fontsize{10}{10}{\textbf{\textcolor{black}{0.44}}}&\text{\textcolor{black}{0.09}} / \text{\textcolor{black}{0.64}}&\text{\textcolor{black}{0.11}} / \text{\textcolor{black}{0.17}}&\text{\textcolor{black}{0.15}} / \text{\textcolor{black}{0.00}}&\text{\textcolor{black}{0.12}} / \fontsize{10}{10}{\textbf{\textcolor{black}{0.00}}}&\fontsize{10}{10}{\textbf{\textcolor{black}{0.09}}} / \fontsize{10}{10}{\textbf{\textcolor{black}{0.07}}}&\text{\textcolor{black}{0.05}} / \fontsize{10}{10}{\textbf{\textcolor{black}{0.54}}}

\\

 L-TOPK (naive) &\fontsize{10}{10}{\textbf{\textcolor{black}{0.12}}} / \text{\textcolor{black}{0.05}}&\text{\textcolor{black}{0.12}} / \text{\textcolor{black}{0.17}}&\text{\textcolor{black}{0.09}} / \text{\textcolor{black}{0.36}}&\text{\textcolor{black}{0.09}} / \text{\textcolor{black}{0.63}}&\text{\textcolor{black}{0.11}} / \text{\textcolor{black}{0.17}}&\text{\textcolor{black}{0.16}} / \text{\textcolor{black}{0.00}}&\fontsize{10}{10}{\textbf{\textcolor{black}{0.11}}} / \text{\textcolor{black}{0.00}}&\text{\textcolor{black}{0.09}} / \text{\textcolor{black}{0.06}}&\text{\textcolor{black}{0.05}} / \text{\textcolor{black}{0.54}}

\\

 \cellcolor{pink}L-TOPK &\cellcolor{pink}\text{\textcolor{black}{0.12}} / \cellcolor{pink}\fontsize{10}{10}{\textbf{\textcolor{black}{0.07}}}&\cellcolor{pink}\fontsize{10}{10}{\textbf{\textcolor{black}{0.12}}} / \cellcolor{pink}\fontsize{10}{10}{\textbf{\textcolor{black}{0.21}}}&\cellcolor{pink}\fontsize{10}{10}{\textbf{\textcolor{black}{0.08}}} / \cellcolor{pink}\text{\textcolor{black}{0.41}}&\cellcolor{pink}\fontsize{10}{10}{\textbf{\textcolor{black}{0.07}}} / \cellcolor{pink}\fontsize{10}{10}{\textbf{\textcolor{black}{0.76}}}&\cellcolor{pink}\fontsize{10}{10}{\textbf{\textcolor{black}{0.10}}} / \cellcolor{pink}\fontsize{10}{10}{\textbf{\textcolor{black}{0.18}}}&\cellcolor{pink}\fontsize{10}{10}{\textbf{\textcolor{black}{0.15}}} / \cellcolor{pink}\fontsize{10}{10}{\textbf{\textcolor{black}{0.02}}}&\cellcolor{pink}\text{\textcolor{black}{0.12}} / \cellcolor{pink}\text{\textcolor{black}{0.00}}&\cellcolor{pink}\text{\textcolor{black}{0.09}} / \cellcolor{pink}\text{\textcolor{black}{0.06}}&\cellcolor{pink}\fontsize{10}{10}{\textbf{\textcolor{black}{0.05}}} / \cellcolor{pink}\text{\textcolor{black}{0.51}}

\\

\midrule

 LAC &\text{\textcolor{black}{0.12}} / \text{\textcolor{black}{0.01}}&\text{\textcolor{black}{0.13}} / \text{\textcolor{black}{0.10}}&\text{\textcolor{black}{0.09}} / \text{\textcolor{black}{0.11}}&\text{\textcolor{black}{0.07}} / \text{\textcolor{black}{0.74}}&\text{\textcolor{black}{0.11}} / \text{\textcolor{black}{0.09}}&\text{\textcolor{black}{0.16}} / \text{\textcolor{black}{0.00}}&\text{\textcolor{black}{0.13}} / \text{\textcolor{black}{0.00}}&\text{\textcolor{black}{0.09}} / \text{\textcolor{black}{0.06}}&\text{\textcolor{black}{0.05}} / \text{\textcolor{black}{0.44}}

\\

 L-LAC (naive) &\fontsize{10}{10}{\textbf{\textcolor{black}{0.11}}} / \text{\textcolor{black}{0.02}}&\text{\textcolor{black}{0.13}} / \text{\textcolor{black}{0.14}}&\text{\textcolor{black}{0.09}} / \text{\textcolor{black}{0.13}}&\text{\textcolor{black}{0.07}} / \text{\textcolor{black}{0.75}}&\text{\textcolor{black}{0.11}} / \fontsize{10}{10}{\textbf{\textcolor{black}{0.12}}}&\text{\textcolor{black}{0.16}} / \text{\textcolor{black}{0.00}}&\text{\textcolor{black}{0.13}} / \text{\textcolor{black}{0.00}}&\fontsize{10}{10}{\textbf{\textcolor{black}{0.09}}} / \fontsize{10}{10}{\textbf{\textcolor{black}{0.07}}}&\fontsize{10}{10}{\textbf{\textcolor{black}{0.05}}} / \fontsize{10}{10}{\textbf{\textcolor{black}{0.51}}}

\\

 \cellcolor{pink}L-LAC &\cellcolor{pink}\text{\textcolor{black}{0.12}} / \cellcolor{pink}\fontsize{10}{10}{\textbf{\textcolor{black}{0.02}}}&\cellcolor{pink}\fontsize{10}{10}{\textbf{\textcolor{black}{0.12}}} / \cellcolor{pink}\fontsize{10}{10}{\textbf{\textcolor{black}{0.19}}}&\cellcolor{pink}\fontsize{10}{10}{\textbf{\textcolor{black}{0.08}}} / \cellcolor{pink}\fontsize{10}{10}{\textbf{\textcolor{black}{0.29}}}&\cellcolor{pink}\fontsize{10}{10}{\textbf{\textcolor{black}{0.07}}} / \cellcolor{pink}\fontsize{10}{10}{\textbf{\textcolor{black}{0.76}}}&\cellcolor{pink}\fontsize{10}{10}{\textbf{\textcolor{black}{0.10}}} / \cellcolor{pink}\text{\textcolor{black}{0.10}}&\cellcolor{pink}\fontsize{10}{10}{\textbf{\textcolor{black}{0.16}}} / \cellcolor{pink}\fontsize{10}{10}{\textbf{\textcolor{black}{0.01}}}&\cellcolor{pink}\fontsize{10}{10}{\textbf{\textcolor{black}{0.12}}} / \cellcolor{pink}\fontsize{10}{10}{\textbf{\textcolor{black}{0.00}}}&\cellcolor{pink}\text{\textcolor{black}{0.09}} / \cellcolor{pink}\text{\textcolor{black}{0.06}}&\cellcolor{pink}\text{\textcolor{black}{0.05}} / \cellcolor{pink}\text{\textcolor{black}{0.45}}

\\

\midrule

 APS &\text{\textcolor{black}{0.11}} / \text{\textcolor{black}{0.05}}&\text{\textcolor{black}{0.12}} / \text{\textcolor{black}{0.14}}&\text{\textcolor{black}{0.08}} / \text{\textcolor{black}{0.90}}&\text{\textcolor{black}{0.06}} / \fontsize{10}{10}{\textbf{\textcolor{black}{0.82}}}&\text{\textcolor{black}{0.09}} / \text{\textcolor{black}{0.21}}&\text{\textcolor{black}{0.15}} / \text{\textcolor{black}{0.00}}&\text{\textcolor{black}{0.13}} / \fontsize{10}{10}{\textbf{\textcolor{black}{0.00}}}&\text{\textcolor{black}{0.08}} / \text{\textcolor{black}{0.21}}&\fontsize{10}{10}{\textbf{\textcolor{black}{0.07}}} / \text{\textcolor{black}{0.72}}

\\

 L-APS (naive) &\text{\textcolor{black}{0.11}} / \text{\textcolor{black}{0.05}}&\text{\textcolor{black}{0.12}} / \text{\textcolor{black}{0.17}}&\text{\textcolor{black}{0.08}} / \fontsize{10}{10}{\textbf{\textcolor{black}{0.91}}}&\text{\textcolor{black}{0.06}} / \text{\textcolor{black}{0.82}}&\text{\textcolor{black}{0.09}} / \fontsize{10}{10}{\textbf{\textcolor{black}{0.26}}}&\text{\textcolor{black}{0.15}} / \text{\textcolor{black}{0.00}}&\text{\textcolor{black}{0.12}} / \text{\textcolor{black}{0.00}}&\text{\textcolor{black}{0.08}} / \fontsize{10}{10}{\textbf{\textcolor{black}{0.22}}}&\text{\textcolor{black}{0.07}} / \fontsize{10}{10}{\textbf{\textcolor{black}{0.75}}}

\\

 \cellcolor{pink}L-APS &\cellcolor{pink}\fontsize{10}{10}{\textbf{\textcolor{black}{0.11}}} / \cellcolor{pink}\fontsize{10}{10}{\textbf{\textcolor{black}{0.05}}}&\cellcolor{pink}\fontsize{10}{10}{\textbf{\textcolor{black}{0.12}}} / \cellcolor{pink}\fontsize{10}{10}{\textbf{\textcolor{black}{0.19}}}&\cellcolor{pink}\fontsize{10}{10}{\textbf{\textcolor{black}{0.08}}} / \cellcolor{pink}\text{\textcolor{black}{0.90}}&\cellcolor{pink}\fontsize{10}{10}{\textbf{\textcolor{black}{0.06}}} / \cellcolor{pink}\text{\textcolor{black}{0.82}}&\cellcolor{pink}\fontsize{10}{10}{\textbf{\textcolor{black}{0.09}}} / \cellcolor{pink}\text{\textcolor{black}{0.23}}&\cellcolor{pink}\fontsize{10}{10}{\textbf{\textcolor{black}{0.14}}} / \cellcolor{pink}\fontsize{10}{10}{\textbf{\textcolor{black}{0.00}}}&\cellcolor{pink}\fontsize{10}{10}{\textbf{\textcolor{black}{0.12}}} / \cellcolor{pink}\text{\textcolor{black}{0.00}}&\cellcolor{pink}\fontsize{10}{10}{\textbf{\textcolor{black}{0.08}}} / \cellcolor{pink}\text{\textcolor{black}{0.21}}&\cellcolor{pink}\text{\textcolor{black}{0.07}} / \cellcolor{pink}\text{\textcolor{black}{0.72}}

\\

\midrule

 RAPS &\text{\textcolor{black}{0.11}} / \text{\textcolor{black}{0.04}}&\text{\textcolor{black}{0.13}} / \text{\textcolor{black}{0.12}}&\text{\textcolor{black}{0.09}} / \text{\textcolor{black}{0.69}}&\text{\textcolor{black}{0.07}} / \fontsize{10}{10}{\textbf{\textcolor{black}{0.81}}}&\text{\textcolor{black}{0.10}} / \text{\textcolor{black}{0.27}}&\fontsize{10}{10}{\textbf{\textcolor{black}{0.16}}} / \text{\textcolor{black}{0.00}}&\text{\textcolor{black}{0.13}} / \fontsize{10}{10}{\textbf{\textcolor{black}{0.00}}}&\text{\textcolor{black}{0.08}} / \fontsize{10}{10}{\textbf{\textcolor{black}{0.08}}}&\text{\textcolor{black}{0.06}} / \text{\textcolor{black}{0.72}}

\\

 L-RAPS (naive) &\fontsize{10}{10}{\textbf{\textcolor{black}{0.11}}} / \fontsize{10}{10}{\textbf{\textcolor{black}{0.06}}}&\text{\textcolor{black}{0.12}} / \fontsize{10}{10}{\textbf{\textcolor{black}{0.13}}}&\text{\textcolor{black}{0.09}} / \fontsize{10}{10}{\textbf{\textcolor{black}{0.70}}}&\fontsize{10}{10}{\textbf{\textcolor{black}{0.07}}} / \text{\textcolor{black}{0.80}}&\text{\textcolor{black}{0.10}} / \fontsize{10}{10}{\textbf{\textcolor{black}{0.29}}}&\text{\textcolor{black}{0.16}} / \text{\textcolor{black}{0.00}}&\fontsize{10}{10}{\textbf{\textcolor{black}{0.13}}} / \text{\textcolor{black}{0.00}}&\fontsize{10}{10}{\textbf{\textcolor{black}{0.08}}} / \text{\textcolor{black}{0.08}}&\text{\textcolor{black}{0.06}} / \fontsize{10}{10}{\textbf{\textcolor{black}{0.74}}}

\\

 \cellcolor{pink}L-RAPS &\cellcolor{pink}\text{\textcolor{black}{0.11}} / \cellcolor{pink}\text{\textcolor{black}{0.04}}&\cellcolor{pink}\fontsize{10}{10}{\textbf{\textcolor{black}{0.12}}} / \cellcolor{pink}\text{\textcolor{black}{0.11}}&\cellcolor{pink}\fontsize{10}{10}{\textbf{\textcolor{black}{0.09}}} / \cellcolor{pink}\text{\textcolor{black}{0.70}}&\cellcolor{pink}\text{\textcolor{black}{0.08}} / \cellcolor{pink}\text{\textcolor{black}{0.68}}&\cellcolor{pink}\fontsize{10}{10}{\textbf{\textcolor{black}{0.10}}} / \cellcolor{pink}\text{\textcolor{black}{0.27}}&\cellcolor{pink}\text{\textcolor{black}{0.16}} / \cellcolor{pink}\fontsize{10}{10}{\textbf{\textcolor{black}{0.00}}}&\cellcolor{pink}\text{\textcolor{black}{0.13}} / \cellcolor{pink}\text{\textcolor{black}{0.00}}&\cellcolor{pink}\text{\textcolor{black}{0.08}} / \cellcolor{pink}\text{\textcolor{black}{0.08}}&\cellcolor{pink}\fontsize{10}{10}{\textbf{\textcolor{black}{0.06}}} / \cellcolor{pink}\text{\textcolor{black}{0.72}}

\\

\midrule

\end{tabular}} \end{subtable}

\vspace{0.15cm}

\begin{subtable}{\linewidth}
        \setlength\dashlinedash{0.2pt}
                            \setlength\dashlinegap{1.5pt}
                            \setlength\arrayrulewidth{0.3pt}
                            \renewcommand{\arraystretch}{0.85}
                                \caption{
            Results with ViT-L/14.}
        \centering
        \resizebox{\textwidth}{!}{
            \begin{tabular}{l|ccccccccc}
              \toprule

 & \rotatebox[origin=c]{45}{UCF101} & \rotatebox[origin=c]{45}{DTD} & \rotatebox[origin=c]{45}{Pets} & \rotatebox[origin=c]{45}{EuroSAT} & \rotatebox[origin=c]{45}{StanfordCars} & \rotatebox[origin=c]{45}{Flower102} & \rotatebox[origin=c]{45}{Aircraft} & \rotatebox[origin=c]{45}{SUN397} & \rotatebox[origin=c]{45}{Food101} 

\\ \midrule

\textbf{Zero-Shot Accuracy} & $\text{75.1}$ & $\text{53.4}$ & $\text{93.5}$ & $\text{60.3}$ & $\text{76.9}$ & $\text{79.6}$ & $\text{32.5}$ & $\text{67.7}$ & $\text{90.9}$ \\ \midrule

 TOPK &\fontsize{10}{10}{\textbf{\textcolor{black}{0.11}}} / \text{\textcolor{black}{0.08}}&\text{\textcolor{black}{0.10}} / \text{\textcolor{black}{0.08}}&\fontsize{10}{10}{\textbf{\textcolor{black}{0.08}}} / \fontsize{10}{10}{\textbf{\textcolor{black}{0.47}}}&\fontsize{10}{10}{\textbf{\textcolor{black}{0.08}}} / \fontsize{10}{10}{\textbf{\textcolor{black}{0.78}}}&\text{\textcolor{black}{0.11}} / \fontsize{10}{10}{\textbf{\textcolor{black}{0.16}}}&\text{\textcolor{black}{0.14}} / \text{\textcolor{black}{0.00}}&\text{\textcolor{black}{0.08}} / \text{\textcolor{black}{0.21}}&\fontsize{10}{10}{\textbf{\textcolor{black}{0.09}}} / \fontsize{10}{10}{\textbf{\textcolor{black}{0.07}}}&\text{\textcolor{black}{0.05}} / \fontsize{10}{10}{\textbf{\textcolor{black}{0.62}}}

\\

 L-TOPK (naive) &\text{\textcolor{black}{0.11}} / \text{\textcolor{black}{0.06}}&\fontsize{10}{10}{\textbf{\textcolor{black}{0.10}}} / \text{\textcolor{black}{0.12}}&\text{\textcolor{black}{0.08}} / \text{\textcolor{black}{0.47}}&\text{\textcolor{black}{0.09}} / \text{\textcolor{black}{0.70}}&\fontsize{10}{10}{\textbf{\textcolor{black}{0.11}}} / \text{\textcolor{black}{0.16}}&\text{\textcolor{black}{0.15}} / \text{\textcolor{black}{0.00}}&\fontsize{10}{10}{\textbf{\textcolor{black}{0.08}}} / \fontsize{10}{10}{\textbf{\textcolor{black}{0.22}}}&\text{\textcolor{black}{0.09}} / \text{\textcolor{black}{0.06}}&\text{\textcolor{black}{0.05}} / \text{\textcolor{black}{0.62}}

\\

 \cellcolor{pink}L-TOPK &\cellcolor{pink}\text{\textcolor{black}{0.11}} / \cellcolor{pink}\fontsize{10}{10}{\textbf{\textcolor{black}{0.10}}}&\cellcolor{pink}\text{\textcolor{black}{0.10}} / \cellcolor{pink}\fontsize{10}{10}{\textbf{\textcolor{black}{0.21}}}&\cellcolor{pink}\text{\textcolor{black}{0.08}} / \cellcolor{pink}\text{\textcolor{black}{0.47}}&\cellcolor{pink}\text{\textcolor{black}{0.08}} / \cellcolor{pink}\text{\textcolor{black}{0.74}}&\cellcolor{pink}\text{\textcolor{black}{0.11}} / \cellcolor{pink}\text{\textcolor{black}{0.13}}&\cellcolor{pink}\fontsize{10}{10}{\textbf{\textcolor{black}{0.14}}} / \cellcolor{pink}\fontsize{10}{10}{\textbf{\textcolor{black}{0.00}}}&\cellcolor{pink}\text{\textcolor{black}{0.08}} / \cellcolor{pink}\text{\textcolor{black}{0.20}}&\cellcolor{pink}\text{\textcolor{black}{0.09}} / \cellcolor{pink}\text{\textcolor{black}{0.05}}&\cellcolor{pink}\fontsize{10}{10}{\textbf{\textcolor{black}{0.05}}} / \cellcolor{pink}\text{\textcolor{black}{0.54}}

\\

\midrule

 LAC &\text{\textcolor{black}{0.12}} / \text{\textcolor{black}{0.01}}&\text{\textcolor{black}{0.10}} / \text{\textcolor{black}{0.04}}&\text{\textcolor{black}{0.09}} / \text{\textcolor{black}{0.24}}&\text{\textcolor{black}{0.08}} / \text{\textcolor{black}{0.68}}&\text{\textcolor{black}{0.11}} / \text{\textcolor{black}{0.06}}&\text{\textcolor{black}{0.15}} / \text{\textcolor{black}{0.00}}&\text{\textcolor{black}{0.09}} / \text{\textcolor{black}{0.16}}&\text{\textcolor{black}{0.09}} / \text{\textcolor{black}{0.13}}&\text{\textcolor{black}{0.05}} / \text{\textcolor{black}{0.49}}

\\

 L-LAC (naive) &\fontsize{10}{10}{\textbf{\textcolor{black}{0.11}}} / \text{\textcolor{black}{0.02}}&\fontsize{10}{10}{\textbf{\textcolor{black}{0.10}}} / \text{\textcolor{black}{0.07}}&\text{\textcolor{black}{0.09}} / \text{\textcolor{black}{0.28}}&\text{\textcolor{black}{0.08}} / \text{\textcolor{black}{0.67}}&\text{\textcolor{black}{0.10}} / \text{\textcolor{black}{0.10}}&\text{\textcolor{black}{0.15}} / \text{\textcolor{black}{0.00}}&\text{\textcolor{black}{0.09}} / \text{\textcolor{black}{0.18}}&\fontsize{10}{10}{\textbf{\textcolor{black}{0.09}}} / \fontsize{10}{10}{\textbf{\textcolor{black}{0.14}}}&\fontsize{10}{10}{\textbf{\textcolor{black}{0.05}}} / \fontsize{10}{10}{\textbf{\textcolor{black}{0.58}}}

\\

 \cellcolor{pink}L-LAC &\cellcolor{pink}\text{\textcolor{black}{0.12}} / \cellcolor{pink}\fontsize{10}{10}{\textbf{\textcolor{black}{0.05}}}&\cellcolor{pink}\text{\textcolor{black}{0.10}} / \cellcolor{pink}\fontsize{10}{10}{\textbf{\textcolor{black}{0.07}}}&\cellcolor{pink}\fontsize{10}{10}{\textbf{\textcolor{black}{0.08}}} / \cellcolor{pink}\fontsize{10}{10}{\textbf{\textcolor{black}{0.32}}}&\cellcolor{pink}\fontsize{10}{10}{\textbf{\textcolor{black}{0.07}}} / \cellcolor{pink}\fontsize{10}{10}{\textbf{\textcolor{black}{0.73}}}&\cellcolor{pink}\fontsize{10}{10}{\textbf{\textcolor{black}{0.10}}} / \cellcolor{pink}\fontsize{10}{10}{\textbf{\textcolor{black}{0.11}}}&\cellcolor{pink}\fontsize{10}{10}{\textbf{\textcolor{black}{0.14}}} / \cellcolor{pink}\fontsize{10}{10}{\textbf{\textcolor{black}{0.00}}}&\cellcolor{pink}\fontsize{10}{10}{\textbf{\textcolor{black}{0.09}}} / \cellcolor{pink}\fontsize{10}{10}{\textbf{\textcolor{black}{0.20}}}&\cellcolor{pink}\text{\textcolor{black}{0.09}} / \cellcolor{pink}\text{\textcolor{black}{0.13}}&\cellcolor{pink}\text{\textcolor{black}{0.05}} / \cellcolor{pink}\text{\textcolor{black}{0.50}}

\\

\midrule

 APS &\text{\textcolor{black}{0.10}} / \text{\textcolor{black}{0.10}}&\text{\textcolor{black}{0.09}} / \text{\textcolor{black}{0.07}}&\text{\textcolor{black}{0.09}} / \text{\textcolor{black}{0.92}}&\text{\textcolor{black}{0.07}} / \text{\textcolor{black}{0.85}}&\text{\textcolor{black}{0.09}} / \text{\textcolor{black}{0.42}}&\text{\textcolor{black}{0.13}} / \fontsize{10}{10}{\textbf{\textcolor{black}{0.00}}}&\text{\textcolor{black}{0.09}} / \text{\textcolor{black}{0.10}}&\text{\textcolor{black}{0.08}} / \text{\textcolor{black}{0.27}}&\fontsize{10}{10}{\textbf{\textcolor{black}{0.07}}} / \text{\textcolor{black}{0.90}}

\\

 L-APS (naive) &\text{\textcolor{black}{0.10}} / \text{\textcolor{black}{0.10}}&\fontsize{10}{10}{\textbf{\textcolor{black}{0.09}}} / \text{\textcolor{black}{0.09}}&\text{\textcolor{black}{0.09}} / \fontsize{10}{10}{\textbf{\textcolor{black}{0.92}}}&\text{\textcolor{black}{0.07}} / \text{\textcolor{black}{0.84}}&\text{\textcolor{black}{0.09}} / \fontsize{10}{10}{\textbf{\textcolor{black}{0.45}}}&\text{\textcolor{black}{0.13}} / \text{\textcolor{black}{0.00}}&\text{\textcolor{black}{0.09}} / \text{\textcolor{black}{0.13}}&\text{\textcolor{black}{0.08}} / \text{\textcolor{black}{0.28}}&\text{\textcolor{black}{0.08}} / \fontsize{10}{10}{\textbf{\textcolor{black}{0.92}}}

\\

 \cellcolor{pink}L-APS &\cellcolor{pink}\fontsize{10}{10}{\textbf{\textcolor{black}{0.10}}} / \cellcolor{pink}\fontsize{10}{10}{\textbf{\textcolor{black}{0.12}}}&\cellcolor{pink}\text{\textcolor{black}{0.09}} / \cellcolor{pink}\fontsize{10}{10}{\textbf{\textcolor{black}{0.11}}}&\cellcolor{pink}\fontsize{10}{10}{\textbf{\textcolor{black}{0.09}}} / \cellcolor{pink}\text{\textcolor{black}{0.92}}&\cellcolor{pink}\fontsize{10}{10}{\textbf{\textcolor{black}{0.07}}} / \cellcolor{pink}\fontsize{10}{10}{\textbf{\textcolor{black}{0.86}}}&\cellcolor{pink}\fontsize{10}{10}{\textbf{\textcolor{black}{0.09}}} / \cellcolor{pink}\text{\textcolor{black}{0.43}}&\cellcolor{pink}\fontsize{10}{10}{\textbf{\textcolor{black}{0.12}}} / \cellcolor{pink}\text{\textcolor{black}{0.00}}&\cellcolor{pink}\fontsize{10}{10}{\textbf{\textcolor{black}{0.09}}} / \cellcolor{pink}\fontsize{10}{10}{\textbf{\textcolor{black}{0.15}}}&\cellcolor{pink}\fontsize{10}{10}{\textbf{\textcolor{black}{0.08}}} / \cellcolor{pink}\fontsize{10}{10}{\textbf{\textcolor{black}{0.29}}}&\cellcolor{pink}\text{\textcolor{black}{0.07}} / \cellcolor{pink}\text{\textcolor{black}{0.90}}

\\

\midrule

 RAPS &\text{\textcolor{black}{0.10}} / \text{\textcolor{black}{0.13}}&\text{\textcolor{black}{0.10}} / \text{\textcolor{black}{0.05}}&\fontsize{10}{10}{\textbf{\textcolor{black}{0.08}}} / \text{\textcolor{black}{0.88}}&\text{\textcolor{black}{0.07}} / \fontsize{10}{10}{\textbf{\textcolor{black}{0.84}}}&\text{\textcolor{black}{0.09}} / \text{\textcolor{black}{0.27}}&\text{\textcolor{black}{0.13}} / \fontsize{10}{10}{\textbf{\textcolor{black}{0.00}}}&\text{\textcolor{black}{0.09}} / \text{\textcolor{black}{0.11}}&\text{\textcolor{black}{0.08}} / \fontsize{10}{10}{\textbf{\textcolor{black}{0.15}}}&\fontsize{10}{10}{\textbf{\textcolor{black}{0.06}}} / \text{\textcolor{black}{0.88}}

\\

 L-RAPS (naive) &\fontsize{10}{10}{\textbf{\textcolor{black}{0.10}}} / \fontsize{10}{10}{\textbf{\textcolor{black}{0.13}}}&\fontsize{10}{10}{\textbf{\textcolor{black}{0.10}}} / \text{\textcolor{black}{0.06}}&\text{\textcolor{black}{0.08}} / \fontsize{10}{10}{\textbf{\textcolor{black}{0.88}}}&\text{\textcolor{black}{0.07}} / \text{\textcolor{black}{0.84}}&\text{\textcolor{black}{0.10}} / \fontsize{10}{10}{\textbf{\textcolor{black}{0.28}}}&\text{\textcolor{black}{0.13}} / \text{\textcolor{black}{0.00}}&\fontsize{10}{10}{\textbf{\textcolor{black}{0.09}}} / \fontsize{10}{10}{\textbf{\textcolor{black}{0.14}}}&\fontsize{10}{10}{\textbf{\textcolor{black}{0.08}}} / \text{\textcolor{black}{0.15}}&\text{\textcolor{black}{0.07}} / \fontsize{10}{10}{\textbf{\textcolor{black}{0.89}}}

\\

 \cellcolor{pink}L-RAPS &\cellcolor{pink}\text{\textcolor{black}{0.10}} / \cellcolor{pink}\text{\textcolor{black}{0.10}}&\cellcolor{pink}\text{\textcolor{black}{0.11}} / \cellcolor{pink}\fontsize{10}{10}{\textbf{\textcolor{black}{0.08}}}&\cellcolor{pink}\text{\textcolor{black}{0.08}} / \cellcolor{pink}\text{\textcolor{black}{0.88}}&\cellcolor{pink}\fontsize{10}{10}{\textbf{\textcolor{black}{0.07}}} / \cellcolor{pink}\text{\textcolor{black}{0.78}}&\cellcolor{pink}\fontsize{10}{10}{\textbf{\textcolor{black}{0.09}}} / \cellcolor{pink}\text{\textcolor{black}{0.28}}&\cellcolor{pink}\fontsize{10}{10}{\textbf{\textcolor{black}{0.13}}} / \cellcolor{pink}\text{\textcolor{black}{0.00}}&\cellcolor{pink}\text{\textcolor{black}{0.09}} / \cellcolor{pink}\text{\textcolor{black}{0.13}}&\cellcolor{pink}\text{\textcolor{black}{0.08}} / \cellcolor{pink}\text{\textcolor{black}{0.15}}&\cellcolor{pink}\text{\textcolor{black}{0.06}} / \cellcolor{pink}\text{\textcolor{black}{0.88}}

\\

\midrule

\end{tabular}} \end{subtable}

\vspace{0.15cm}

 \begin{subtable}{\linewidth}
        \setlength\dashlinedash{0.2pt}
                            \setlength\dashlinegap{1.5pt}
                            \setlength\arrayrulewidth{0.3pt}
                            \renewcommand{\arraystretch}{0.85}
                                \caption{
            Results with RN50.}
        \centering
        \resizebox{\textwidth}{!}{
            \begin{tabular}{l|ccccccccc}
              \toprule

 & \rotatebox[origin=c]{45}{UCF101} & \rotatebox[origin=c]{45}{DTD} & \rotatebox[origin=c]{45}{Pets} & \rotatebox[origin=c]{45}{EuroSAT} & \rotatebox[origin=c]{45}{StanfordCars} & \rotatebox[origin=c]{45}{Flower102} & \rotatebox[origin=c]{45}{Aircraft} & \rotatebox[origin=c]{45}{SUN397} & \rotatebox[origin=c]{45}{Food101} 

\\ \midrule

\textbf{Zero-Shot Accuracy} & $\text{61.9}$ & $\text{42.8}$ & $\text{85.7}$ & $\text{36.1}$ & $\text{55.8}$ & $\text{66.0}$ & $\text{17.0}$ & $\text{58.8}$ & $\text{77.4}$ \\ \midrule

 TOPK &\text{\textcolor{black}{0.12}} / \text{\textcolor{black}{0.02}}&\text{\textcolor{black}{0.12}} / \text{\textcolor{black}{0.03}}&\text{\textcolor{black}{0.07}} / \fontsize{10}{10}{\textbf{\textcolor{black}{0.48}}}&\fontsize{10}{10}{\textbf{\textcolor{black}{0.08}}} / \fontsize{10}{10}{\textbf{\textcolor{black}{0.81}}}&\text{\textcolor{black}{0.10}} / \fontsize{10}{10}{\textbf{\textcolor{black}{0.08}}}&\text{\textcolor{black}{0.14}} / \text{\textcolor{black}{0.00}}&\text{\textcolor{black}{0.11}} / \text{\textcolor{black}{0.00}}&\fontsize{10}{10}{\textbf{\textcolor{black}{0.08}}} / \fontsize{10}{10}{\textbf{\textcolor{black}{0.03}}}&\fontsize{10}{10}{\textbf{\textcolor{black}{0.05}}} / \fontsize{10}{10}{\textbf{\textcolor{black}{0.70}}}

\\

 L-TOPK (naive) &\fontsize{10}{10}{\textbf{\textcolor{black}{0.11}}} / \fontsize{10}{10}{\textbf{\textcolor{black}{0.03}}}&\text{\textcolor{black}{0.12}} / \text{\textcolor{black}{0.04}}&\text{\textcolor{black}{0.07}} / \text{\textcolor{black}{0.48}}&\text{\textcolor{black}{0.11}} / \text{\textcolor{black}{0.68}}&\fontsize{10}{10}{\textbf{\textcolor{black}{0.10}}} / \text{\textcolor{black}{0.08}}&\text{\textcolor{black}{0.15}} / \text{\textcolor{black}{0.00}}&\text{\textcolor{black}{0.11}} / \text{\textcolor{black}{0.00}}&\text{\textcolor{black}{0.08}} / \text{\textcolor{black}{0.03}}&\text{\textcolor{black}{0.05}} / \text{\textcolor{black}{0.70}}

\\

 \cellcolor{pink}L-TOPK &\cellcolor{pink}\text{\textcolor{black}{0.12}} / \cellcolor{pink}\text{\textcolor{black}{0.02}}&\cellcolor{pink}\fontsize{10}{10}{\textbf{\textcolor{black}{0.11}}} / \cellcolor{pink}\fontsize{10}{10}{\textbf{\textcolor{black}{0.14}}}&\cellcolor{pink}\fontsize{10}{10}{\textbf{\textcolor{black}{0.06}}} / \cellcolor{pink}\text{\textcolor{black}{0.48}}&\cellcolor{pink}\text{\textcolor{black}{0.08}} / \cellcolor{pink}\text{\textcolor{black}{0.76}}&\cellcolor{pink}\text{\textcolor{black}{0.10}} / \cellcolor{pink}\text{\textcolor{black}{0.08}}&\cellcolor{pink}\fontsize{10}{10}{\textbf{\textcolor{black}{0.14}}} / \cellcolor{pink}\fontsize{10}{10}{\textbf{\textcolor{black}{0.01}}}&\cellcolor{pink}\fontsize{10}{10}{\textbf{\textcolor{black}{0.10}}} / \cellcolor{pink}\fontsize{10}{10}{\textbf{\textcolor{black}{0.00}}}&\cellcolor{pink}\text{\textcolor{black}{0.08}} / \cellcolor{pink}\text{\textcolor{black}{0.03}}&\cellcolor{pink}\text{\textcolor{black}{0.05}} / \cellcolor{pink}\text{\textcolor{black}{0.68}}

\\

\midrule

 LAC &\text{\textcolor{black}{0.12}} / \text{\textcolor{black}{0.02}}&\text{\textcolor{black}{0.12}} / \text{\textcolor{black}{0.09}}&\text{\textcolor{black}{0.07}} / \text{\textcolor{black}{0.35}}&\text{\textcolor{black}{0.09}} / \text{\textcolor{black}{0.70}}&\text{\textcolor{black}{0.10}} / \text{\textcolor{black}{0.08}}&\text{\textcolor{black}{0.14}} / \fontsize{10}{10}{\textbf{\textcolor{black}{0.00}}}&\text{\textcolor{black}{0.11}} / \text{\textcolor{black}{0.01}}&\text{\textcolor{black}{0.08}} / \text{\textcolor{black}{0.02}}&\text{\textcolor{black}{0.05}} / \text{\textcolor{black}{0.66}}

\\

 L-LAC (naive) &\fontsize{10}{10}{\textbf{\textcolor{black}{0.11}}} / \text{\textcolor{black}{0.03}}&\text{\textcolor{black}{0.11}} / \text{\textcolor{black}{0.13}}&\text{\textcolor{black}{0.07}} / \text{\textcolor{black}{0.39}}&\text{\textcolor{black}{0.09}} / \text{\textcolor{black}{0.68}}&\text{\textcolor{black}{0.10}} / \fontsize{10}{10}{\textbf{\textcolor{black}{0.09}}}&\text{\textcolor{black}{0.15}} / \text{\textcolor{black}{0.00}}&\text{\textcolor{black}{0.11}} / \text{\textcolor{black}{0.01}}&\text{\textcolor{black}{0.08}} / \text{\textcolor{black}{0.03}}&\fontsize{10}{10}{\textbf{\textcolor{black}{0.05}}} / \fontsize{10}{10}{\textbf{\textcolor{black}{0.70}}}

\\

 \cellcolor{pink}L-LAC &\cellcolor{pink}\text{\textcolor{black}{0.12}} / \cellcolor{pink}\fontsize{10}{10}{\textbf{\textcolor{black}{0.03}}}&\cellcolor{pink}\fontsize{10}{10}{\textbf{\textcolor{black}{0.11}}} / \cellcolor{pink}\fontsize{10}{10}{\textbf{\textcolor{black}{0.19}}}&\cellcolor{pink}\fontsize{10}{10}{\textbf{\textcolor{black}{0.07}}} / \cellcolor{pink}\fontsize{10}{10}{\textbf{\textcolor{black}{0.42}}}&\cellcolor{pink}\fontsize{10}{10}{\textbf{\textcolor{black}{0.07}}} / \cellcolor{pink}\fontsize{10}{10}{\textbf{\textcolor{black}{0.75}}}&\cellcolor{pink}\fontsize{10}{10}{\textbf{\textcolor{black}{0.10}}} / \cellcolor{pink}\text{\textcolor{black}{0.08}}&\cellcolor{pink}\fontsize{10}{10}{\textbf{\textcolor{black}{0.14}}} / \cellcolor{pink}\text{\textcolor{black}{0.00}}&\cellcolor{pink}\fontsize{10}{10}{\textbf{\textcolor{black}{0.11}}} / \cellcolor{pink}\fontsize{10}{10}{\textbf{\textcolor{black}{0.01}}}&\cellcolor{pink}\fontsize{10}{10}{\textbf{\textcolor{black}{0.08}}} / \cellcolor{pink}\fontsize{10}{10}{\textbf{\textcolor{black}{0.03}}}&\cellcolor{pink}\text{\textcolor{black}{0.05}} / \cellcolor{pink}\text{\textcolor{black}{0.66}}

\\

\midrule

 APS &\text{\textcolor{black}{0.11}} / \text{\textcolor{black}{0.08}}&\text{\textcolor{black}{0.12}} / \text{\textcolor{black}{0.10}}&\text{\textcolor{black}{0.07}} / \text{\textcolor{black}{0.90}}&\text{\textcolor{black}{0.06}} / \fontsize{10}{10}{\textbf{\textcolor{black}{0.83}}}&\text{\textcolor{black}{0.09}} / \text{\textcolor{black}{0.19}}&\text{\textcolor{black}{0.14}} / \text{\textcolor{black}{0.00}}&\text{\textcolor{black}{0.11}} / \text{\textcolor{black}{0.00}}&\text{\textcolor{black}{0.08}} / \text{\textcolor{black}{0.09}}&\text{\textcolor{black}{0.05}} / \text{\textcolor{black}{0.81}}

\\

 L-APS (naive) &\fontsize{10}{10}{\textbf{\textcolor{black}{0.11}}} / \fontsize{10}{10}{\textbf{\textcolor{black}{0.08}}}&\fontsize{10}{10}{\textbf{\textcolor{black}{0.11}}} / \text{\textcolor{black}{0.13}}&\text{\textcolor{black}{0.07}} / \fontsize{10}{10}{\textbf{\textcolor{black}{0.90}}}&\text{\textcolor{black}{0.06}} / \text{\textcolor{black}{0.81}}&\text{\textcolor{black}{0.09}} / \fontsize{10}{10}{\textbf{\textcolor{black}{0.21}}}&\text{\textcolor{black}{0.14}} / \text{\textcolor{black}{0.00}}&\fontsize{10}{10}{\textbf{\textcolor{black}{0.11}}} / \fontsize{10}{10}{\textbf{\textcolor{black}{0.01}}}&\text{\textcolor{black}{0.08}} / \text{\textcolor{black}{0.09}}&\text{\textcolor{black}{0.06}} / \fontsize{10}{10}{\textbf{\textcolor{black}{0.83}}}

\\

 \cellcolor{pink}L-APS &\cellcolor{pink}\text{\textcolor{black}{0.11}} / \cellcolor{pink}\text{\textcolor{black}{0.08}}&\cellcolor{pink}\text{\textcolor{black}{0.11}} / \cellcolor{pink}\fontsize{10}{10}{\textbf{\textcolor{black}{0.18}}}&\cellcolor{pink}\fontsize{10}{10}{\textbf{\textcolor{black}{0.07}}} / \cellcolor{pink}\text{\textcolor{black}{0.90}}&\cellcolor{pink}\fontsize{10}{10}{\textbf{\textcolor{black}{0.06}}} / \cellcolor{pink}\text{\textcolor{black}{0.83}}&\cellcolor{pink}\fontsize{10}{10}{\textbf{\textcolor{black}{0.09}}} / \cellcolor{pink}\text{\textcolor{black}{0.19}}&\cellcolor{pink}\fontsize{10}{10}{\textbf{\textcolor{black}{0.14}}} / \cellcolor{pink}\fontsize{10}{10}{\textbf{\textcolor{black}{0.01}}}&\cellcolor{pink}\text{\textcolor{black}{0.11}} / \cellcolor{pink}\text{\textcolor{black}{0.00}}&\cellcolor{pink}\fontsize{10}{10}{\textbf{\textcolor{black}{0.08}}} / \cellcolor{pink}\fontsize{10}{10}{\textbf{\textcolor{black}{0.09}}}&\cellcolor{pink}\fontsize{10}{10}{\textbf{\textcolor{black}{0.05}}} / \cellcolor{pink}\text{\textcolor{black}{0.81}}

\\

\midrule

 RAPS &\text{\textcolor{black}{0.11}} / \text{\textcolor{black}{0.04}}&\text{\textcolor{black}{0.12}} / \text{\textcolor{black}{0.05}}&\text{\textcolor{black}{0.07}} / \text{\textcolor{black}{0.81}}&\fontsize{10}{10}{\textbf{\textcolor{black}{0.08}}} / \fontsize{10}{10}{\textbf{\textcolor{black}{0.72}}}&\text{\textcolor{black}{0.09}} / \text{\textcolor{black}{0.12}}&\text{\textcolor{black}{0.15}} / \fontsize{10}{10}{\textbf{\textcolor{black}{0.00}}}&\text{\textcolor{black}{0.11}} / \text{\textcolor{black}{0.00}}&\text{\textcolor{black}{0.08}} / \fontsize{10}{10}{\textbf{\textcolor{black}{0.03}}}&\text{\textcolor{black}{0.05}} / \text{\textcolor{black}{0.75}}

\\

 L-RAPS (naive) &\fontsize{10}{10}{\textbf{\textcolor{black}{0.11}}} / \fontsize{10}{10}{\textbf{\textcolor{black}{0.04}}}&\fontsize{10}{10}{\textbf{\textcolor{black}{0.12}}} / \text{\textcolor{black}{0.06}}&\text{\textcolor{black}{0.08}} / \text{\textcolor{black}{0.81}}&\text{\textcolor{black}{0.09}} / \text{\textcolor{black}{0.71}}&\fontsize{10}{10}{\textbf{\textcolor{black}{0.09}}} / \fontsize{10}{10}{\textbf{\textcolor{black}{0.12}}}&\text{\textcolor{black}{0.15}} / \text{\textcolor{black}{0.00}}&\fontsize{10}{10}{\textbf{\textcolor{black}{0.11}}} / \text{\textcolor{black}{0.00}}&\text{\textcolor{black}{0.08}} / \text{\textcolor{black}{0.03}}&\text{\textcolor{black}{0.05}} / \fontsize{10}{10}{\textbf{\textcolor{black}{0.76}}}

\\

 \cellcolor{pink}L-RAPS &\cellcolor{pink}\text{\textcolor{black}{0.11}} / \cellcolor{pink}\text{\textcolor{black}{0.04}}&\cellcolor{pink}\text{\textcolor{black}{0.12}} / \cellcolor{pink}\fontsize{10}{10}{\textbf{\textcolor{black}{0.06}}}&\cellcolor{pink}\fontsize{10}{10}{\textbf{\textcolor{black}{0.07}}} / \cellcolor{pink}\fontsize{10}{10}{\textbf{\textcolor{black}{0.81}}}&\cellcolor{pink}\text{\textcolor{black}{0.12}} / \cellcolor{pink}\text{\textcolor{black}{0.52}}&\cellcolor{pink}\text{\textcolor{black}{0.09}} / \cellcolor{pink}\text{\textcolor{black}{0.11}}&\cellcolor{pink}\fontsize{10}{10}{\textbf{\textcolor{black}{0.15}}} / \cellcolor{pink}\text{\textcolor{black}{0.00}}&\cellcolor{pink}\text{\textcolor{black}{0.12}} / \cellcolor{pink}\fontsize{10}{10}{\textbf{\textcolor{black}{0.00}}}&\cellcolor{pink}\fontsize{10}{10}{\textbf{\textcolor{black}{0.08}}} / \cellcolor{pink}\text{\textcolor{black}{0.03}}&\cellcolor{pink}\fontsize{10}{10}{\textbf{\textcolor{black}{0.05}}} / \cellcolor{pink}\text{\textcolor{black}{0.75}}

\\

\midrule

\end{tabular}} \end{subtable}

\vspace{0.15cm}

\begin{subtable}{\linewidth}
        \setlength\dashlinedash{0.2pt}
                            \setlength\dashlinegap{1.5pt}
                            \setlength\arrayrulewidth{0.3pt}
                            \renewcommand{\arraystretch}{0.85}
                                \caption{
            Results with RN101.}
        \label{tab:results}
        \centering
        \resizebox{\textwidth}{!}{
            \begin{tabular}{l|ccccccccc}
              \toprule

 & \rotatebox[origin=c]{45}{UCF101} & \rotatebox[origin=c]{45}{DTD} & \rotatebox[origin=c]{45}{Pets} & \rotatebox[origin=c]{45}{EuroSAT} & \rotatebox[origin=c]{45}{StanfordCars} & \rotatebox[origin=c]{45}{Flower102} & \rotatebox[origin=c]{45}{Aircraft} & \rotatebox[origin=c]{45}{SUN397} & \rotatebox[origin=c]{45}{Food101} 

\\ \midrule

\textbf{Zero-Shot Accuracy} & $\text{61.0}$ & $\text{37.1}$ & $\text{86.9}$ & $\text{32.8}$ & $\text{63.2}$ & $\text{64.4}$ & $\text{18.1}$ & $\text{59.0}$ & $\text{80.7}$ \\ \midrule

 TOPK &\text{\textcolor{black}{0.12}} / \text{\textcolor{black}{0.00}}&\text{\textcolor{black}{0.11}} / \text{\textcolor{black}{0.07}}&\text{\textcolor{black}{0.09}} / \text{\textcolor{black}{0.11}}&\text{\textcolor{black}{0.11}} / \text{\textcolor{black}{0.61}}&\text{\textcolor{black}{0.10}} / \fontsize{10}{10}{\textbf{\textcolor{black}{0.00}}}&\text{\textcolor{black}{0.14}} / \text{\textcolor{black}{0.00}}&\fontsize{10}{10}{\textbf{\textcolor{black}{0.11}}} / \text{\textcolor{black}{0.00}}&\text{\textcolor{black}{0.09}} / \text{\textcolor{black}{0.01}}&\text{\textcolor{black}{0.05}} / \fontsize{10}{10}{\textbf{\textcolor{black}{0.52}}}

\\

 L-TOPK (naive) &\fontsize{10}{10}{\textbf{\textcolor{black}{0.12}}} / \text{\textcolor{black}{0.00}}&\text{\textcolor{black}{0.11}} / \text{\textcolor{black}{0.06}}&\text{\textcolor{black}{0.12}} / \text{\textcolor{black}{0.03}}&\text{\textcolor{black}{0.13}} / \text{\textcolor{black}{0.48}}&\text{\textcolor{black}{0.10}} / \text{\textcolor{black}{0.00}}&\text{\textcolor{black}{0.15}} / \text{\textcolor{black}{0.00}}&\text{\textcolor{black}{0.12}} / \text{\textcolor{black}{0.00}}&\fontsize{10}{10}{\textbf{\textcolor{black}{0.08}}} / \fontsize{10}{10}{\textbf{\textcolor{black}{0.02}}}&\text{\textcolor{black}{0.06}} / \text{\textcolor{black}{0.50}}

\\

 \cellcolor{pink}L-TOPK &\cellcolor{pink}\text{\textcolor{black}{0.12}} / \cellcolor{pink}\fontsize{10}{10}{\textbf{\textcolor{black}{0.00}}}&\cellcolor{pink}\fontsize{10}{10}{\textbf{\textcolor{black}{0.11}}} / \cellcolor{pink}\fontsize{10}{10}{\textbf{\textcolor{black}{0.19}}}&\cellcolor{pink}\fontsize{10}{10}{\textbf{\textcolor{black}{0.08}}} / \cellcolor{pink}\fontsize{10}{10}{\textbf{\textcolor{black}{0.13}}}&\cellcolor{pink}\fontsize{10}{10}{\textbf{\textcolor{black}{0.09}}} / \cellcolor{pink}\fontsize{10}{10}{\textbf{\textcolor{black}{0.67}}}&\cellcolor{pink}\fontsize{10}{10}{\textbf{\textcolor{black}{0.09}}} / \cellcolor{pink}\text{\textcolor{black}{0.00}}&\cellcolor{pink}\fontsize{10}{10}{\textbf{\textcolor{black}{0.14}}} / \cellcolor{pink}\fontsize{10}{10}{\textbf{\textcolor{black}{0.01}}}&\cellcolor{pink}\text{\textcolor{black}{0.12}} / \cellcolor{pink}\fontsize{10}{10}{\textbf{\textcolor{black}{0.01}}}&\cellcolor{pink}\text{\textcolor{black}{0.09}} / \cellcolor{pink}\text{\textcolor{black}{0.01}}&\cellcolor{pink}\fontsize{10}{10}{\textbf{\textcolor{black}{0.05}}} / \cellcolor{pink}\text{\textcolor{black}{0.51}}

\\

\midrule

 LAC &\text{\textcolor{black}{0.12}} / \text{\textcolor{black}{0.01}}&\text{\textcolor{black}{0.11}} / \text{\textcolor{black}{0.15}}&\text{\textcolor{black}{0.09}} / \text{\textcolor{black}{0.04}}&\text{\textcolor{black}{0.10}} / \fontsize{10}{10}{\textbf{\textcolor{black}{0.64}}}&\text{\textcolor{black}{0.10}} / \text{\textcolor{black}{0.00}}&\text{\textcolor{black}{0.15}} / \text{\textcolor{black}{0.00}}&\text{\textcolor{black}{0.13}} / \text{\textcolor{black}{0.00}}&\text{\textcolor{black}{0.09}} / \text{\textcolor{black}{0.01}}&\text{\textcolor{black}{0.05}} / \text{\textcolor{black}{0.50}}

\\

 L-LAC (naive) &\fontsize{10}{10}{\textbf{\textcolor{black}{0.11}}} / \text{\textcolor{black}{0.01}}&\text{\textcolor{black}{0.11}} / \text{\textcolor{black}{0.14}}&\text{\textcolor{black}{0.09}} / \text{\textcolor{black}{0.03}}&\text{\textcolor{black}{0.10}} / \text{\textcolor{black}{0.62}}&\text{\textcolor{black}{0.10}} / \text{\textcolor{black}{0.00}}&\text{\textcolor{black}{0.15}} / \text{\textcolor{black}{0.00}}&\text{\textcolor{black}{0.13}} / \text{\textcolor{black}{0.00}}&\fontsize{10}{10}{\textbf{\textcolor{black}{0.08}}} / \fontsize{10}{10}{\textbf{\textcolor{black}{0.02}}}&\text{\textcolor{black}{0.05}} / \text{\textcolor{black}{0.51}}

\\

 \cellcolor{pink}L-LAC &\cellcolor{pink}\text{\textcolor{black}{0.12}} / \cellcolor{pink}\fontsize{10}{10}{\textbf{\textcolor{black}{0.01}}}&\cellcolor{pink}\fontsize{10}{10}{\textbf{\textcolor{black}{0.11}}} / \cellcolor{pink}\fontsize{10}{10}{\textbf{\textcolor{black}{0.22}}}&\cellcolor{pink}\fontsize{10}{10}{\textbf{\textcolor{black}{0.08}}} / \cellcolor{pink}\fontsize{10}{10}{\textbf{\textcolor{black}{0.15}}}&\cellcolor{pink}\fontsize{10}{10}{\textbf{\textcolor{black}{0.09}}} / \cellcolor{pink}\text{\textcolor{black}{0.63}}&\cellcolor{pink}\fontsize{10}{10}{\textbf{\textcolor{black}{0.10}}} / \cellcolor{pink}\fontsize{10}{10}{\textbf{\textcolor{black}{0.01}}}&\cellcolor{pink}\fontsize{10}{10}{\textbf{\textcolor{black}{0.13}}} / \cellcolor{pink}\fontsize{10}{10}{\textbf{\textcolor{black}{0.03}}}&\cellcolor{pink}\fontsize{10}{10}{\textbf{\textcolor{black}{0.12}}} / \cellcolor{pink}\fontsize{10}{10}{\textbf{\textcolor{black}{0.01}}}&\cellcolor{pink}\text{\textcolor{black}{0.09}} / \cellcolor{pink}\text{\textcolor{black}{0.01}}&\cellcolor{pink}\fontsize{10}{10}{\textbf{\textcolor{black}{0.05}}} / \cellcolor{pink}\fontsize{10}{10}{\textbf{\textcolor{black}{0.53}}}

\\

\midrule

 APS &\text{\textcolor{black}{0.10}} / \text{\textcolor{black}{0.02}}&\text{\textcolor{black}{0.11}} / \text{\textcolor{black}{0.19}}&\text{\textcolor{black}{0.09}} / \text{\textcolor{black}{0.64}}&\text{\textcolor{black}{0.09}} / \fontsize{10}{10}{\textbf{\textcolor{black}{0.73}}}&\text{\textcolor{black}{0.09}} / \fontsize{10}{10}{\textbf{\textcolor{black}{0.00}}}&\text{\textcolor{black}{0.14}} / \text{\textcolor{black}{0.02}}&\text{\textcolor{black}{0.13}} / \text{\textcolor{black}{0.00}}&\text{\textcolor{black}{0.08}} / \text{\textcolor{black}{0.07}}&\fontsize{10}{10}{\textbf{\textcolor{black}{0.06}}} / \text{\textcolor{black}{0.75}}

\\

 L-APS (naive) &\text{\textcolor{black}{0.10}} / \fontsize{10}{10}{\textbf{\textcolor{black}{0.03}}}&\text{\textcolor{black}{0.11}} / \text{\textcolor{black}{0.18}}&\text{\textcolor{black}{0.09}} / \text{\textcolor{black}{0.62}}&\text{\textcolor{black}{0.09}} / \text{\textcolor{black}{0.72}}&\text{\textcolor{black}{0.09}} / \text{\textcolor{black}{0.00}}&\text{\textcolor{black}{0.14}} / \text{\textcolor{black}{0.01}}&\text{\textcolor{black}{0.13}} / \text{\textcolor{black}{0.00}}&\text{\textcolor{black}{0.08}} / \fontsize{10}{10}{\textbf{\textcolor{black}{0.09}}}&\text{\textcolor{black}{0.06}} / \fontsize{10}{10}{\textbf{\textcolor{black}{0.76}}}

\\

 \cellcolor{pink}L-APS &\cellcolor{pink}\fontsize{10}{10}{\textbf{\textcolor{black}{0.10}}} / \cellcolor{pink}\text{\textcolor{black}{0.03}}&\cellcolor{pink}\fontsize{10}{10}{\textbf{\textcolor{black}{0.11}}} / \cellcolor{pink}\fontsize{10}{10}{\textbf{\textcolor{black}{0.28}}}&\cellcolor{pink}\fontsize{10}{10}{\textbf{\textcolor{black}{0.09}}} / \cellcolor{pink}\fontsize{10}{10}{\textbf{\textcolor{black}{0.68}}}&\cellcolor{pink}\fontsize{10}{10}{\textbf{\textcolor{black}{0.08}}} / \cellcolor{pink}\text{\textcolor{black}{0.72}}&\cellcolor{pink}\fontsize{10}{10}{\textbf{\textcolor{black}{0.09}}} / \cellcolor{pink}\text{\textcolor{black}{0.00}}&\cellcolor{pink}\fontsize{10}{10}{\textbf{\textcolor{black}{0.13}}} / \cellcolor{pink}\fontsize{10}{10}{\textbf{\textcolor{black}{0.03}}}&\cellcolor{pink}\fontsize{10}{10}{\textbf{\textcolor{black}{0.12}}} / \cellcolor{pink}\fontsize{10}{10}{\textbf{\textcolor{black}{0.00}}}&\cellcolor{pink}\fontsize{10}{10}{\textbf{\textcolor{black}{0.08}}} / \cellcolor{pink}\text{\textcolor{black}{0.07}}&\cellcolor{pink}\text{\textcolor{black}{0.06}} / \cellcolor{pink}\text{\textcolor{black}{0.75}}

\\

\midrule

 RAPS &\text{\textcolor{black}{0.12}} / \fontsize{10}{10}{\textbf{\textcolor{black}{0.00}}}&\text{\textcolor{black}{0.12}} / \text{\textcolor{black}{0.08}}&\text{\textcolor{black}{0.09}} / \text{\textcolor{black}{0.37}}&\fontsize{10}{10}{\textbf{\textcolor{black}{0.15}}} / \fontsize{10}{10}{\textbf{\textcolor{black}{0.45}}}&\text{\textcolor{black}{0.09}} / \fontsize{10}{10}{\textbf{\textcolor{black}{0.00}}}&\text{\textcolor{black}{0.14}} / \fontsize{10}{10}{\textbf{\textcolor{black}{0.00}}}&\fontsize{10}{10}{\textbf{\textcolor{black}{0.13}}} / \fontsize{10}{10}{\textbf{\textcolor{black}{0.00}}}&\fontsize{10}{10}{\textbf{\textcolor{black}{0.08}}} / \text{\textcolor{black}{0.02}}&\text{\textcolor{black}{0.06}} / \text{\textcolor{black}{0.61}}

\\

 L-RAPS (naive) &\fontsize{10}{10}{\textbf{\textcolor{black}{0.11}}} / \text{\textcolor{black}{0.00}}&\fontsize{10}{10}{\textbf{\textcolor{black}{0.12}}} / \text{\textcolor{black}{0.08}}&\text{\textcolor{black}{0.09}} / \text{\textcolor{black}{0.37}}&\text{\textcolor{black}{0.15}} / \text{\textcolor{black}{0.43}}&\text{\textcolor{black}{0.09}} / \text{\textcolor{black}{0.00}}&\text{\textcolor{black}{0.15}} / \text{\textcolor{black}{0.00}}&\text{\textcolor{black}{0.13}} / \text{\textcolor{black}{0.00}}&\text{\textcolor{black}{0.08}} / \fontsize{10}{10}{\textbf{\textcolor{black}{0.02}}}&\text{\textcolor{black}{0.06}} / \fontsize{10}{10}{\textbf{\textcolor{black}{0.61}}}

\\

 \cellcolor{pink}L-RAPS &\cellcolor{pink}\text{\textcolor{black}{0.12}} / \cellcolor{pink}\text{\textcolor{black}{0.00}}&\cellcolor{pink}\text{\textcolor{black}{0.12}} / \cellcolor{pink}\fontsize{10}{10}{\textbf{\textcolor{black}{0.11}}}&\cellcolor{pink}\fontsize{10}{10}{\textbf{\textcolor{black}{0.09}}} / \cellcolor{pink}\fontsize{10}{10}{\textbf{\textcolor{black}{0.38}}}&\cellcolor{pink}\text{\textcolor{black}{0.18}} / \cellcolor{pink}\text{\textcolor{black}{0.39}}&\cellcolor{pink}\fontsize{10}{10}{\textbf{\textcolor{black}{0.09}}} / \cellcolor{pink}\text{\textcolor{black}{0.00}}&\cellcolor{pink}\fontsize{10}{10}{\textbf{\textcolor{black}{0.14}}} / \cellcolor{pink}\text{\textcolor{black}{0.00}}&\cellcolor{pink}\text{\textcolor{black}{0.14}} / \cellcolor{pink}\text{\textcolor{black}{0.00}}&\cellcolor{pink}\text{\textcolor{black}{0.08}} / \cellcolor{pink}\text{\textcolor{black}{0.02}}&\cellcolor{pink}\fontsize{10}{10}{\textbf{\textcolor{black}{0.06}}} / \cellcolor{pink}\text{\textcolor{black}{0.61}}

\\

\midrule

\end{tabular}} \end{subtable}

\end{table*}

\section{Conclusion}\label{sec:conclusion}
In this paper, we presented an extensive benchmarking of localized conformal prediction on a variety of image classification tasks with several VLMs and conformal scores. We showed how the naive approach, using a cosine similarity in the visual latent space (an intuitive choice for VLMs), fails to improve over the non-local baselines, even degrading the mean set size in several cases. We thus introduced a sigmoidal transformation of the cosine similarities, using cross-validation on the calibration samples to tune its hyperparameters. This approach allowed to consistently improve over the non-local baselines across datasets and backbones, often achieving a statistically significant reduction of set sizes. Conversely, we showed that improvements in classical coverage-related metrics were less impacted by the local algorithms, with the coverage gap very closely following the behavior of the non-local baselines. We hope these findings as well as our open source code will contribute to the deployment of localized conformal prediction algorithms in real-world applications.
\bibliographystyle{IEEEtranS.bst}
\bibliography{icme2025references}
\clearpage
\let\oldthetable\thetable
\let\oldthefigure\thefigure

\renewcommand{\thetable}{A-\Roman{table}}
\renewcommand{\thefigure}{A-\Roman{figure}}

\setcounter{table}{0} 
\setcounter{figure}{0}
\renewcommand{\thesection}{\Alph{section}}
\setcounter{section}{0}
\setcounter{page}{1}
\onecolumn
\begin{center}
    {\Large \textbf{Localized Conformal Prediction for Image Classification with Vision-Language Models}}\\[10pt]
    {\Large Supplementary Materials}\\[5pt]
\end{center}

\vspace{1em}

\section{Dataset details}
\label{sec:supp_datasets}

\begin{table}[h]
    \centering
    \caption{Additional information on the datasets.}
    \label{tab:appendix_datasets}
    \resizebox{0.75\linewidth}{!}{
    \begin{tabular}{lccc}
    \toprule
    Dataset name & Classes & Samples & Task \\
    \midrule
    UCF101 & 101 & 13320 & Action classification \\
    DTD & 47 & 5640 & Texture classification \\
    Pets & 37 & 7349 & Pet breed classification \\
    EuroSAT & 10 & 27000 & Satellite image classification \\
    StanfordCars & 196 & 16185 & Car model classification \\
    Flower102 & 102 & 8189 & Flower classification \\
    Aircraft & 100 & 10000 & Aircraft model classification \\
    SUN397 & 397 & 39700 & Scene classification \\
    Food101 & 101 & 101000 & Food classification \\
    \bottomrule
    \end{tabular}}
\end{table}
\begin{figure}[h]
\begin{center}
\includegraphics[width=\textwidth]{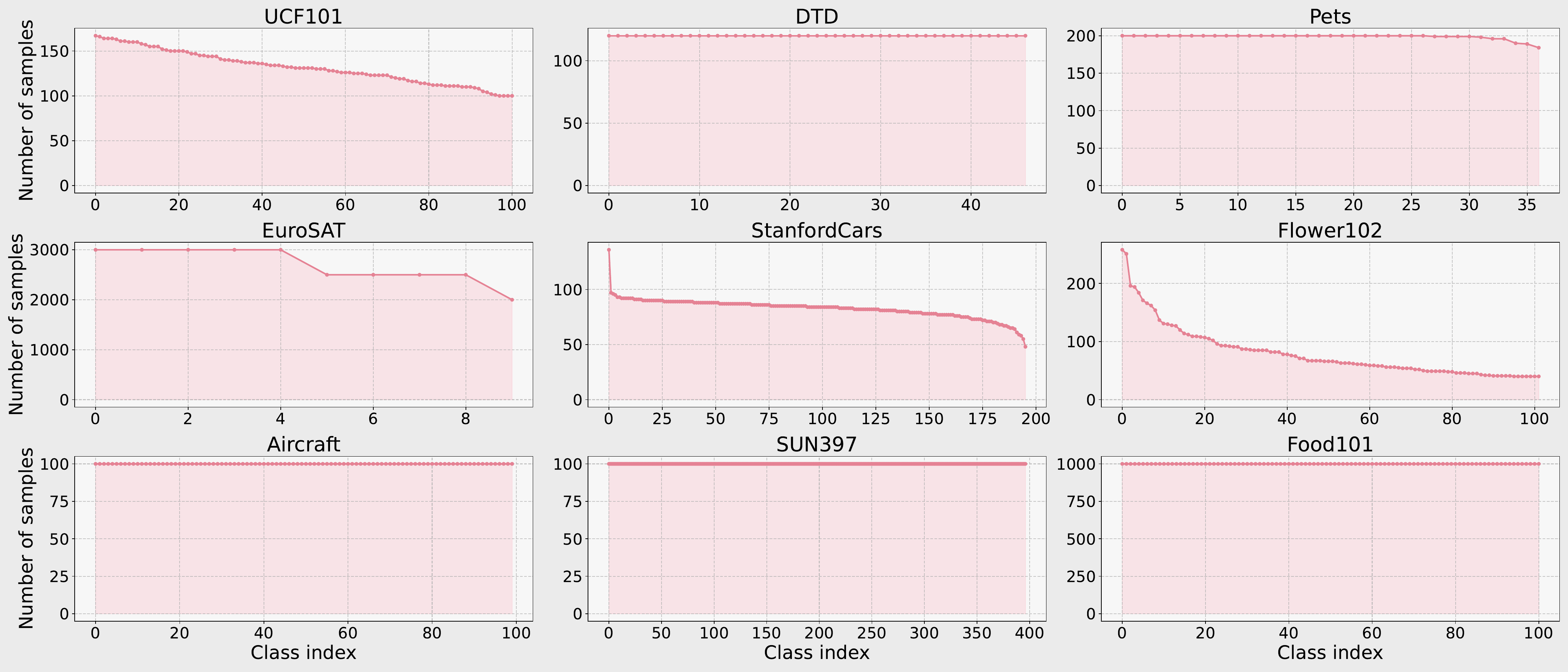}
\end{center}
\caption{Labels distribution for each dataset. Note that our calibration sets are uniformly drawn from the dataset, therefore conserving the same distribution.}\label{fig:datasets}
\vspace{-10pt}
\end{figure}

\end{document}